\documentclass[11pt]{article}
\usepackage{array}
\usepackage[tbtags]{amsmath}
\usepackage{graphicx}
\usepackage{amsfonts}
\usepackage{amssymb}
\usepackage{mathrsfs}
\usepackage{indentfirst}
\usepackage{dsfont}
\usepackage{subfigure}
\usepackage{caption}
\usepackage {float}
\usepackage{epstopdf}
\usepackage{multirow}
\usepackage{booktabs}
\usepackage[colorlinks, linkcolor=blue, anchorcolor=blue, citecolor=blue]{hyperref}
\usepackage[authoryear,round]{natbib}
\usepackage[ruled,vlined]{algorithm2e}

\usepackage{tikz}
\usetikzlibrary{arrows.meta,positioning,calc,fit,backgrounds}
\usepackage{geometry}
\usepackage{framed}
\usepackage[normalem]{ulem}
\usepackage{calc}
\usepackage{xcolor}
\usetikzlibrary{calc}
\usetikzlibrary{positioning}
\usetikzlibrary{topaths}
\usetikzlibrary{automata}
\usepackage{enumitem}

\providecommand{\U}[1]{\protect\rule{.1in}{.1in}}
\newtheorem{theorem}{Theorem}[section]

\newtheorem{assumption}{Assumption}[section]

\newtheorem{corollary}{Corollary}[section]

\newtheorem{lemma}{Lemma}[section]

\newtheorem{remark}{Remark}[section]

\def\E{\mathbb{E}}

\input epsf

\date{}

\def\bSig\mathbf{\Sigma}

\begin{document}
	
\begingroup
\renewcommand{\thefootnote}{\fnsymbol{footnote}}
\footnotetext[1]{Corresponding author, email: bingyijing@cuhk.edu.cn}
\endgroup

\title{\bf Model-based Bootstrap for Offline Policy Evaluation in Tabular Reinforcement Learning
}
\author{Weiwei Wang$^{1}$,\ Yuqiang Li$^{2,*}$,\ Xianyi Wu$^{2,*}$ and Bingyi Jing$^{3,4,}$
\footnote{E-mail addresses: Weiwei Wang, weiweiwang\_stat@163.com; Yuqiang Li, yqli@stat.ecnu.edu.cn; Xianyi Wu, xywu@stat.ecnu.edu.cn}\\
$^{1}$ Department of Statistics and Data Science, Southern University of Science and Technology\\
$^{2}$ School of Statistics \& KLATASDS-MOE, East China Normal University\\
$^{3}$ School of Artificial Intelligence, The Chinese University of Hong Kong, Shenzhen\\
$^{4}$ Shenzhen Loop Area Institute}

\maketitle

\begin{abstract}
Offline policy evaluation (OPE) is crucial in high-stakes reinforcement learning applications, where new policies must be assessed reliably before deployment. In such settings, point estimates alone are insufficient; principled uncertainty quantification, such as confidence intervals and variance estimates, is essential for safe and risk-aware decision-making. A comprehensive way to unify these tasks is to estimate the sampling distribution of the evaluation error. Existing approaches, however, often suffer from limited robustness, scalability, or finite-sample validity. In this paper, we propose a model-based bootstrap framework for uncertainty quantification of OPE in finite-horizon, time-inhomogeneous Markov decision processes (MDPs). Unlike classical bootstrap methods that rely on resampling complete episodes, the proposed method regenerates trajectories from an estimated MDP and can therefore accommodate a much broader range of offline data formats, including complete trajectories, transition-level observations, and trajectory fragments. This flexibility further improves finite-sample statistical efficiency. We establish bootstrap distributional consistency, asymptotically valid confidence intervals, and consistent variance estimation for the target policy value. Extensive simulations show that the proposed method accurately captures the sampling distribution of the OPE estimator, yielding tighter confidence intervals and more accurate variance estimates in most settings.

\vskip3mm

\noindent{\it Keywords}: Bootstrap, confidence interval, Markov decision process, offline policy evaluation, reinforcement learning, uncertainty quantification.

\end{abstract}

\vskip3mm

\maketitle

\baselineskip20pt

\section{Introduction}\label{Sec:intro}

Reinforcement learning (RL) addresses sequential decision-making problems that are commonly modeled as Markov decision processes (MDPs), with the goal of learning policies that maximize cumulative rewards \citep{Sutton:2018}. Over recent decades, RL has achieved notable success across a broad range of domains, including healthcare, education, robotics and autonomous driving \citep{Luckett:2020, Riedmann:2025, Han:2023, Li:2026}. In many such applications, however, online interaction with the environment is impractical due to safety, cost, or time constraints. This has fueled the development of offline reinforcement learning (ORL), where policies are learned solely from pre-collected offline datasets and large-scale historical data can be effectively leveraged \citep{Levine:2020}.

Within this paradigm, offline policy evaluation (OPE) is of central importance: before deploying a policy in a safety-critical setting, one must assess not only its expected performance but also the uncertainty associated with that assessment. Most existing OPE studies focus on point estimation, e.g., \citet{Fonteneau:2010}, \citet{Dudik:2011}, \citet{Thomas:2015}, \citet{Thomas:2016}, \citet{Jiang:2016}, \citet{Liu:2018}, \citet{Farajtabar:2018}, \citet{Xie:2019}, \citet{Yin:2020}, \citet{Duan:2020}, \citet{Tang:2023}, \citet{Wang:2024}, \citet{Liu:2025} and \citet{Liu:2026}, among many others. However, point estimates alone are insufficient in practice, where a policy that appears superior may in fact underperform once uncertainty is accounted for. In high-stakes applications, reliable decision-making requires principled uncertainty quantification, such as confidence intervals and variance estimates, in order to distinguish genuinely superior policies from those that only appear favorable due to statistical noise.

Despite substantial progress in offline policy evaluation, high-confidence OPE remains relatively underdeveloped. Early work mainly focused on confidence intervals for the expected policy value. For instance, \citet{Thomas:2015} proposed a high-confidence estimator based on importance sampling (IS) \citep{Precup:2000} and concentration inequalities. However, IS typically requires a known behavior policy and suffers from the ``Curse of Horizon" \citep{Liu:2018}, often resulting in overly conservative bounds. \citet{Hanna:2017} combined Efron’s bootstrap \citep{Efron:1979} with model-based and weighted doubly robust estimators, but without consistency guarantees. \citet{Kostrikov:2020} proposed bootstrapping independent transitions, i.e., $(s,a,r,s')$, and established asymptotic guarantees under sufficient coverage assumptions; however, \citet{Hao:2021} later showed that transition-level resampling may fail to faithfully recover the error distribution in OPE. To address this issue, \citet{Hao:2021} integrated Efron’s bootstrap with fitted Q-evaluation (FQE) and derived theoretical confidence guarantees. \citet{Shi:2021} further developed a deeply debiased procedure for constructing asymptotic confidence intervals under smoothness assumptions on the Q-function. Another line of work formulates interval estimation as an optimization problem \citep{Feng:2020,Feng:2021,Dai:2020}. While effective in specific settings, these methods are tailored to confidence interval construction and do not directly yield a distributionally consistent approximation that can be reused for broader inferential tasks, such as variance estimation.

More recently, \citet{Shi:2024} studied confidence interval estimation for infinite-horizon off-policy evaluation in the presence of unobserved confounding. \citet{Dann:2023} investigated simultaneous high-confidence evaluation of multiple target policies in the on-policy setting. \citet{Luo:2026} introduced the first simultaneous inference framework for off-policy evaluation, extending pointwise confidence intervals to uniformly valid simultaneous confidence regions over continuous or infinite initial-state spaces. In parallel, conformal approaches have recently been developed for contextual bandits, finite-horizon MDPs, and more general sequential decision-making problems under distribution shift \citep{Taufiq:2022,Foffano:2023,Zhang:2023}. While these methods offer attractive finite-sample coverage guarantees, their primary inferential goal is coverage calibration.

In this paper, we study uncertainty quantification for OPE in finite-horizon, time-inhomogeneous tabular MDPs. Our goal is to go beyond point estimation and develop a trajectory-regeneration bootstrap framework for both confidence interval construction and variance estimation. To this end, we propose a model-based bootstrap (MB) procedure that regenerates trajectories from an estimated MDP and consistently approximates the sampling distribution of the policy value estimator while preserving the Markovian structure of the data. Unlike the classical episode bootstrap, which fundamentally relies on resampling complete observed trajectories, the proposed procedure operates at the level of the estimated MDP and can therefore exploit a much broader class of offline data, including complete trajectories, transition-level observations, and trajectory fragments. This substantially broadens the scope of bootstrap-based inference and makes the proposed framework particularly appealing for data-constrained and partially logged offline RL settings. Our main contributions are summarized as follows.
\begin{itemize}
\item[(1)]  We establish conditional distributional consistency of the proposed bootstrap procedure for both Monte Carlo estimation in the on-policy case and Plug-in estimation (a minimax-optimal OPE method) in the off-policy case. These results provide a mathematically rigorous justification for model-based bootstrap inference in offline RL and fill an important theoretical gap in the current literature, where uncertainty quantification remains much less developed than point estimation.
\item[(2)] Building on this distributional theory, we show that the proposed procedure yields asymptotically valid bootstrap confidence intervals and consistent variance estimation. Hence, the framework provides a unified statistical inference framework that goes substantially beyond point estimation, enabling principled uncertainty quantification for offline policy evaluation.
\item[(3)] We conduct simulation studies in two representative tabular RL environments: Time-varying MDP (a nonstationary MDP with a fixed horizon) and the Cliff-walking environment (a stationary MDP with a random horizon). The experiments assess the accuracy of the proposed model-based bootstrap in approximating the sampling distribution, constructing confidence intervals, and estimating variance. The results show that the proposed method faithfully captures the error distribution, achieves a favorable balance between coverage accuracy and interval efficiency, and provides accurate variance estimation in most settings.
\end{itemize}

The remainder of the paper is organized as follows. Section~\ref{Sec:2} introduces the model setup and reviews several commonly used OPE estimators. Section~\ref{Sec:3} presents the model-based bootstrap procedure, develops its theoretical properties, and describes the construction of confidence intervals and variance estimators in both on-policy and off-policy settings. Section~\ref{Sec:4} reports simulation results, and Section~\ref{Sec:5} concludes the paper. All proofs are deferred to the Appendix.

\section{Preliminaries}\label{Sec:2}

Let $\Delta(\mathcal{S})$ denote the set of all probability distributions over a set $\mathcal{S}$ and for a positive integer $H$, define $[H] := \{0,1,\dots,H-1\}$. We consider a finite-horizon, time-inhomogeneous MDP $\mathcal{M} = (\mathcal{S}, \mathcal{A}, \left\{P_h\right\}_{h \in [H]}, \left\{R_h\right\}_{h \in [H]}, d_0, H)$, where $\mathcal{S}$ and $\mathcal{A}$ are finite state and action spaces, $d_0 \in  \Delta(\mathcal{S})$ is the initial state distribution, and $H<\infty$ is the step horizon. For each step $h \in [H]$, $P_h(\cdot|s,a) \in \Delta(\mathcal{S})$ stands for the transition kernel, and $R_h:\mathcal{S}\times\mathcal{A}\times\mathcal{S} \rightarrow \left[0,1\right]$ indicates the reward function with $r_h(s,a,s')$ the immediate random reward received upon the transition $(s,a) \mapsto s'$ at step $h$. 

A policy $\pi  = {\left\{\pi _h \right\}}_{h \in [H]}$ is a collection of decision rules with $\pi _h(\cdot|s) \in \Delta(\mathcal{A})$. A trajectory under a policy $\pi$ is denoted by $\xi =\left( s_0,a_0,r_0, \ldots ,s_{H-1},a_{H-1},r_{H-1}, s_H \right)$, whose distribution $\mathds{P}^\pi$ is an integration of ${s_0} \sim {d_0}$, $a_h \sim \pi_h \left ( \cdot |s_h \right)$, and $s_{h+1} \sim P_{h} \left(\cdot |s_h, a_h\right)$ for all $h \in \left[ H \right]$. Denote by $G\left( \xi  \right) = \sum_{h = 0}^{H-1} {r_h}$ its return, and $\E_{\mathcal{M}, \pi}$ (or $\E_{\pi}$ if no ambiguity arises) the corresponding expectation. Denote the state- and action-value functions of $\pi$ at step $h$ by 
\begin{equation*}
{V_h^\pi }(s): = \E_{\pi}\left ( \sum\nolimits_{h' = h}^{H-1} {r_{h'}|s_h = s} \right)\hbox{ and }{Q_h^\pi }(s,a): = \E_{\pi}\left ( \sum\nolimits_{h' = h}^{H-1}{r_{h'}|s_h = s,a_h = a}\right),
\end{equation*} respectively, and the overall performance by $v_{\pi} : = \E_{\pi}\left[G\left( \xi \right)\right]$.

\paragraph{Goal.} Given an offline dataset
\begin{equation*}\label{Data_set}
\mathcal{D} = \left\{ \xi ^{(i)} = \left\{ s_0^{(i)}, a_0^{(i)}, r_0^{(i)}, \ldots, s_H^{(i)} \right\} \right\}_{i = 1}^n,
\end{equation*}
of $n$ independent trajectories collected by the target policy $\pi$ or possibly a different behavior policy $\mu = {\left\{\mu_h\right\}_{h \in [H]}}$, we aim to construct: 
\begin{itemize}
\item a confidence interval $\mathrm{CI}(\delta)$ with coverage probability $1 - \delta$, $\delta \in (0,1)$, such that ${\mathds{P}}\left(v_{\pi} \in \mathrm{CI}(\delta)\right) \ge 1-\delta$ and
\item an estimate of the variance of the OPE $\hat v_{\pi}(\mathcal{D})$, i.e., ${\widehat{\mathrm{Var}}}\left(\hat v_{\pi}(\mathcal{D})\right)$.
\end{itemize}

\paragraph{Offline Policy Evaluation.} This study considers the following typical on-policy Monte Carlo (MC) and off-policy Plug-in evaluation methods:
\begin{itemize}
\item {\bf{On-policy MC}} method evaluates a policy by averaging its returns. Specifically, given a data set $\mathcal{D}$, it is defined as
\begin{equation*}
\hat v_{\mathrm{MC}}^{\pi} = \frac{1}{n}\sum_{i=1}^nG(\xi ^{(i)}).
\end{equation*}
The MC estimator is unbiased and consistent. Moreover, by the classical central limit theorem,
\begin{equation}\label{Equ:Asym_distribution:MC}
\sqrt n \left(\hat v_{\mathrm{MC}}^{\pi}  - v_{\pi} \right) \Rightarrow N\left (0, \sigma_{\pi}^2 \right), \quad \text{as} \quad n \to \infty,
\end{equation}
where $\Rightarrow$ denotes converging in distribution, and $\sigma_{\pi}^2:=\mathrm{Var}_{\pi}\left[G(\xi)\right]$, for whose definition please refer to Lemma A.3 in Appendix.
\item {\bf{Off-policy Plug-in}} estimate is defined as
\begin{equation*}
\hat Q_h^\pi(s,a) = \frac{\sum_{i=1}^n \mathbf 1\{s_h^{(i)}=s,\ a_h^{(i)}=a\}\bigl(r_h^{(i)}+\hat V_{h+1}^\pi(s_{h+1}^{(i)})\bigr)}{\sum_{i=1}^n \mathbf 1\{s_h^{(i)}=s,\ a_h^{(i)}=a\}},
\end{equation*}
whenever the denominator is positive, and set $\hat V_H^\pi(s)\equiv 0$ and
\begin{equation*}
\hat V_h^\pi(s)=\sum_{a\in\mathcal A}\pi_h(a\mid s)\hat Q_h^\pi(s,a), 
\end{equation*}
for $h=H-1,\dots,0$. 
The Plug-in value estimator is
\begin{equation}\label{v_plugin}
\hat v_{\mathrm{Plug\text{-}in}}^\pi = \sum_{s\in\mathcal S}d_0(s)\hat V_0^\pi(s).
\end{equation}

\cite{Hao:2021} has proved that $\hat v_{\mathrm{Plug\text{-}in}}^\pi$ is equivalent to the fitted Q-evaluation (FQE), which is known to achieve minimax-optimality in several settings, including tabular MDPs, linear function approximation, and certain nonlinear function classes (e.g., \citet{Duan:2020, Hao:2021, Zhang:2022}). 

\begin{lemma}[Asymptotic normality of the Plug-in estimator] \label{Lem:OPE:asymp_normal}
The Plug-in estimator is $\sqrt{n}$-consistent and asymptotically normal:
\begin{equation*}
\sqrt n\bigl(\hat v_{\mathrm{Plug\text{-}in}}^{\pi} - v_\pi\bigr) \Rightarrow N(0,\sigma_{\mathrm{Plug\text{-}in}}^2).
\end{equation*}
The asymptotic variance is
\begin{equation*}
\sigma_{\mathrm{Plug\text{-}in}}^2 = \sum_{h=0}^{H-1}\mathbb E_\mu\!\left[\left(\frac{d_h^\pi(s_h, a_h)}{d_h^\mu(s_h, a_h)}\right)^2 \operatorname{Var}\!\left(r_h+V_{h+1}^\pi(s_{h+1})\mid s_h, a_h\right)\right],
\end{equation*}
where 
$d_h^\mu(s,a):=\mathbb P_\mu(s_h=s,a_h=a)$.
\end{lemma}

\begin{remark}
The MC is generally statistically inefficient, as its asymptotic variance $\sigma_{\pi}^2$ does not attain the Cram\'er-Rao (C-R) lower bound; see Lemma A.4 in Appendix. Conversely, Lemma~\ref{Lem:OPE:asymp_normal} establishes the asymptotic efficiency of the Plug-in estimator. Relatedly, \cite{Hao:2021} established analogous efficiency results for linear time-homogeneous MDPs. 
\end{remark}
\end{itemize}

\section{Model-based Bootstrap for OPE}\label{Sec:3}

For OPE, in addition to point estimation, it is also critical to quantify uncertainty via, e.g., confidence intervals and variance estimation. A comprehensive way to unify these tasks is to estimate the distribution of the evaluation error $\hat v_{\pi} - v_{\pi}$, denoted by $\mathbb{G}$.  

A standard approach to approximating $\mathbb{G}$ is Efron’s bootstrap. Prior works (e.g., \citet{Hanna:2017, Kostrikov:2020, Hao:2021}) have shown that bootstrap methods can provide asymptotically valid confidence intervals and serve as a promising technique for OPE. However, \citet{Hao:2021}'s numerical experiments demonstrated that resampling should be performed at the trajectory level, because resampling at the transition level ignores the strong dependence between transitions within a trajectory. We refer to these two approaches as bootstrap by episodes (BE) and bootstrap by transitions (BT), respectively.

Specifically, given $\mathcal{D}$, the BE procedure generates a bootstrap dataset $\mathcal{D}^*$ by sampling trajectories with replacement. Applying the MC estimator to $\mathcal{D}^*$ yields the BE-MC estimator $\hat v_{\mathrm{BE\text{-}MC}}^{\pi,*}$. Its distributional consistency
\begin{equation*}
\sqrt n \left (\hat v_{\mathrm{BE\text{-}MC}}^{\pi,*}  - \hat v_{\mathrm{MC}}^{\pi} \right)
\Rightarrow N(0, \sigma_{\pi}^2), 
\end{equation*}
can be readily established by Lemma A.1 in Appendix. Although BE is valid, it is purely nonparametric and may suffer from high variability, especially in small samples. Furthermore, \citet{Hao:2021} proved that the BE-FQE off-policy method (i.e., applying the FQE to $\mathcal{D}^*$) in linear episodic homogeneous MDPs is distributionally consistent, and established the consistency of bootstrap confidence interval as well as bootstrap variance estimation. 

To improve statistical efficiency and relax the complete-trajectory requirement, we propose a \emph{model-based bootstrap} (MB) procedure that leverages the underlying Markovian structure to estimate $\mathbb{G}$ more efficiently. Compared with the nonparametric BE procedure, the proposed MB method can reduce variance and improve finite-sample efficiency, particularly in limited-data settings. We first introduce the model-based bootstrap procedure in Sec.~\ref{Sec:3.1}, and then present the corresponding confidence interval and variance estimation procedures in Sec.~\ref{Sec:3.3}.

\subsection{Model-based bootstrap procedure}\label{Sec:3.1}

Specifically, given the offline dataset $\mathcal D$, we construct the empirical MDP
\begin{equation}\label{MDP:hat}
\widehat{\mathcal M} = \bigl(\mathcal S,\mathcal A,\{\hat P_h\}_{h\in[H]},\{\hat R_h\}_{h\in[H]},d_0,H\bigr),
\end{equation}
where the transition kernel is estimated by
\begin{equation*}\label{Equ:phat}
\hat P_h(s'\mid s,a)=\frac{n_h(s,a,s')}{n_h(s,a)},
\end{equation*}
with $n_h(s,a,s')$ and $n_h(s,a)$ denoting the empirical counts of $(s,a,s')$ and $(s,a)$ at step $h$, respectively. For each $(h,s,a,s')$, let
\begin{equation*}
{\mathcal D}_h(s,a,s') = \Bigl\{(s_h^{(i)},a_h^{(i)},r_h^{(i)},s_{h+1}^{(i)}): (s_h^{(i)},a_h^{(i)},s_{h+1}^{(i)})=(s,a,s'),\ i=1,\ldots,n\Bigr\}
\end{equation*}
be the collection of observed transitions $(s,a) \mapsto s'$ at step $h$, and define $\hat R_h(s,a,s')$ as the sample mean of the rewards in ${\mathcal D}_h(s,a,s')$. Obviously, for every $(h,s,a,s')$ visited with positive probability, the empirical estimators satisfy
\begin{equation*}
\hat P_h(s' \mid s,a) \xrightarrow{p} P_h(s' \mid s,a), \qquad \hat R_h(s,a,s') \xrightarrow{p} R_h(s,a,s'). 
\end{equation*}
Once $\widehat{\mathcal M}$ is available, bootstrap can be performed using transition-level or fragment-level data as well.

For any policy $\bar\pi$, a model-based bootstrap trajectory generated from $(\widehat{\mathcal M},\bar\pi)$ is denoted by
\begin{equation*}
\xi^* = \bigl(s_{0}^*, a_{0}^*, r_{0}^*, \ldots, s_{H-1}^*, a_{H-1}^*, r_{H-1}^*, s_{H}^* \bigr),
\end{equation*}
where $s_{0}^*\sim d_0$, $a_{h}^*\sim \bar\pi_h(\cdot\mid s_{h}^*)$, $s_{h+1}^*\sim \hat P_h(\cdot\mid s_{h}^*, a_{h}^*)$, and $r_{h}^*$ is sampled from the empirical reward distribution associated with the observed transitions in ${\mathcal D}_h\bigl(s_{h}^*, a_{h}^*, s_{h+1}^*\bigr)$. In this way, we generate a bootstrap dataset
\begin{equation*}
\widetilde{\mathcal D}_{\bar\pi}^* = \{\xi^{*(j)}\}_{j=1}^n, \qquad \xi^{*}\sim (\widehat{\mathcal M},\bar\pi),
\end{equation*}
which is then used to conduct statistical inference for the target policy $\pi$. 

\begin{figure}[t]
\centering
\begin{tikzpicture}[x=0.75cm,y=0.75cm]
\tikzset{
panel/.style={
draw=black!70,
rounded corners=8pt,
line width=0.8pt
},
title/.style={
font=\bfseries\footnotesize,
align=center
},
label/.style={
font=\scriptsize
},
note/.style={
font=\footnotesize,
align=center
},
state/.style={
circle,
draw=black!75,
line width=0.85pt,
minimum size=7mm,
inner sep=0pt,
font=\small
},
statenew/.style={
circle,
draw=black,
line width=1.05pt,
minimum size=7mm,
inner sep=0pt,
font=\small
},
arrobs/.style={
-{Latex[length=2.4mm,width=2.0mm]},
draw=black!60,
line width=0.9pt
},
arrmodel/.style={
-{Latex[length=2.5mm,width=2.1mm]},
draw=black!80,
line width=0.95pt
},
arrnew/.style={
-{Latex[length=2.8mm,width=2.3mm]},
draw=black,
line width=1.1pt
},
flow/.style={
-{Latex[length=4mm,width=3.2mm]},
draw=black!80,
line width=1.35pt
}
}

\draw[panel] (0,0) rectangle (5.7,5.6);
\draw[panel] (6.9,0) rectangle (12.6,5.6);
\draw[panel] (13.8,0) rectangle (19.5,5.6);

\node[title] at (2.85,5.22)  {Observed offline data};
\node[title] at (9.75,5.22)  {Pool local transitions};
\node[title] at (16.65,5.22) {Regenerated trajectory};

\node[label, anchor=west] at (0.35,4.6) {Trajectory 1};
\node[state] (L11) at (1.00,3.95) {$s_0$};
\node[state] (L12) at (2.85,3.95) {$s_1$};
\node[state] (L13) at (4.70,3.95) {$s_3$};
\draw[arrobs] (L11) -- (L12);
\draw[arrobs] (L12) -- (L13);

\draw[black!20, dashed, line width=0.7pt] (0.45,3.30) -- (5.25,3.30);

\node[label, anchor=west] at (0.35,2.95) {Trajectory 2};
\node[state] (L21) at (1.00,2.35) {$s_0$};
\node[state] (L22) at (2.85,2.35) {$s_2$};
\node[state] (L23) at (4.70,2.35) {$s_4$};
\draw[arrobs] (L21) -- (L22);
\draw[arrobs] (L22) -- (L23);

\draw[black!20, dashed, line width=0.7pt] (0.45,1.65) -- (5.25,1.65);

\node[label, anchor=west] at (0.35,1.35) {Fragment};
\node[state] (LF1) at (1.00,0.72) {$s_2$};
\node[state] (LF2) at (2.85,0.72) {$s_3$};
\draw[arrobs] (LF1) -- (LF2);

\node[state] (M0) at (9.75,4.10) {$s_0$};
\node[state] (M1) at (8.10,2.95) {$s_1$};
\node[state] (M2) at (11.10,2.95) {$s_2$};
\node[state] (M3) at (8.10,1.25) {$s_3$};
\node[state] (M4) at (11.10,1.25) {$s_4$};

\draw[arrmodel] (M0) -- (M1);
\draw[arrmodel] (M0) -- (M2);
\draw[arrmodel] (M1) -- (M3);
\draw[arrmodel] (M2) -- (M4);
\draw[arrmodel] (M2) -- (M3);

\node[statenew] (R1) at (14.95,3) {$s_0$};
\node[statenew] (R2) at (16.45,3) {$s_2$};
\node[statenew] (R3) at (17.95,3) {$s_3$};

\draw[arrnew] (R1) -- (R2);
\draw[arrnew] (R2) -- (R3);

\node[note] at (16.45,1.80)
{A new trajectory.};

\draw[flow] (5.95,2.55) -- (6.55,2.55);
\draw[flow] (12.85,2.55) -- (13.45,2.55);

\end{tikzpicture}
\caption{Illustration of the efficient use of offline data by the proposed model-based bootstrap. Observed trajectories and fragments are first pooled to estimate the local transition dynamics of the MDP, after which new trajectories can be regenerated from the estimated model. In this example, the regenerated trajectory $s_0 \to s_2 \to s_3$ is supported by local transition information scattered across different samples, even though the full trajectory is not directly observed.}
\label{fig:mb_bootstrap_bw}
\end{figure}
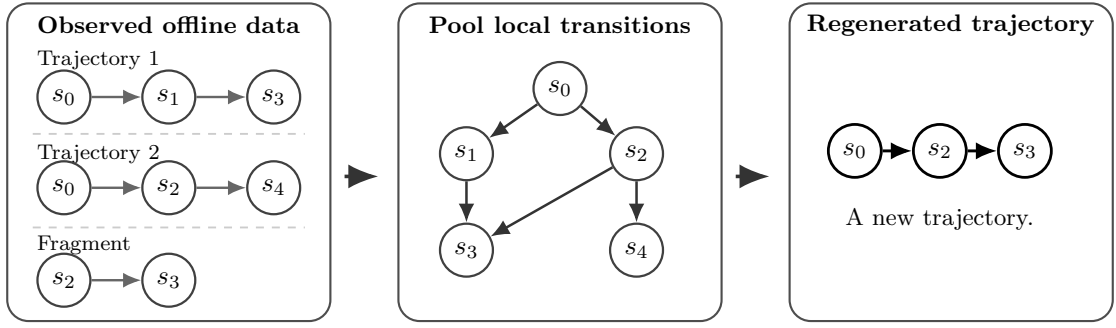

The advantage of regenerating trajectories from an estimated MDP is not merely computational. By consolidating compatible local dynamics across samples, the proposed procedure makes more efficient use of the information contained in the offline dataset than methods that resample complete observed episodes, as illustrated in Fig.~\ref{fig:mb_bootstrap_bw}.

\begin{remark}
The regenerated trajectories considered in this paper are closely connected to the synthetic trajectories in \cite{Wang:2024}, essentially reflecting the same mechanism from a different viewpoint. Specifically, \cite{Wang:2024} breaks observed trajectories into transitions and then re-stitch them on a ``sampling-with-replacement" basis, with the primary goal of improving point estimation accuracy. By contrast, our work adopts the regenerated-trajectory viewpoint to study uncertainty quantification. 
\end{remark}

In what follows, we investigate this idea separately in the on-policy and off-policy settings.

\subsubsection{On-policy bootstrap}\label{Sec:3.1.1}

For on-policy evaluation, the bootstrap dataset $\widetilde{\mathcal D}_{\pi}^*$ is generated under the target policy $\pi$. The corresponding expectation is denoted by
\begin{equation*}
\hat v_{\pi} := \mathbb E_{\widehat{\mathcal M}, \pi}\left[G(\xi^*)\right].
\end{equation*}
Applying the MC estimator to $\widetilde{\mathcal D}_{\pi}^{*}$ yields the model-based bootstrap estimator $\hat v_{\mathrm{MB\text{-}MC}}^{\pi,*}$.

In the sequel, we study the asymptotic properties of the estimator and show how it can be used for CI construction and variance estimation.

\begin{theorem}[On-policy validity of MB-MC]\label{Them:PB:MC:asymp_normal}
Suppose that the dataset $\mathcal D=\{\xi^{(i)}\}_{i=1}^n$ consists of $n$ i.i.d.\ trajectories generated under the target policy $\pi$. Then the following statements hold: as $n \to \infty$,
\begin{enumerate}
\item Expectation consistency:
$$\hat v_{\pi} \xrightarrow{p} v_\pi.$$
\item Variance consistency:
\begin{equation*}
\hat\sigma_\pi^2 \xrightarrow{p} \sigma_\pi^2,\quad \hbox{where} \quad \hat\sigma_\pi^2 := \mathrm{Var}_{\widehat{\mathcal M},\pi}\bigl(G(\xi^*)\bigr)\quad\hbox{and}\quad\sigma_\pi^2 := \mathrm{Var}_{\pi}\bigl(G(\xi)\bigr).
\end{equation*}
\item Conditional asymptotic normality: conditional on $\mathcal D$,
\begin{equation*}
\sqrt n\Bigl(\hat v_{\mathrm{MB\text{-}MC}}^{\pi,*}-\hat v_{\pi}\Bigr)
\Rightarrow
N(0, \sigma_\pi^2).
\end{equation*}
\item Distributional consistency:
\begin{equation*}
\sup_{t\in\mathbb R}\left| \mathbb P\!\left(\sqrt n(\hat v_{\mathrm{MB\text{-}MC}}^{\pi,*}-\hat v_{\pi}) \le t \mid \mathcal D\right) - \mathbb P\!\left(\sqrt n(\hat v_{\mathrm{MC}}^{\pi}-v_\pi)\le t\right)\right|\xrightarrow{p} 0.
\end{equation*}
\end{enumerate}
\end{theorem}

\begin{remark}
Theorem~\ref{Them:PB:MC:asymp_normal} shows that the proposed MB-MC procedure consistently approximates the asymptotic sampling distribution of the MC estimator in the on-policy setting; see (\ref{Equ:Asym_distribution:MC}). Besides, in the absence of distribution shift, the MB correctly reproduces the asymptotic fluctuations of the MC estimator. A key feature is that the bootstrap estimator is centered at the expectation value induced by the estimated MDP, rather than the original MC estimator itself.
\end{remark}

Denote the lower $\delta$-quantile of MB-MC estimate error distribution by
\begin{equation*}
q_{\delta, \mathrm{MB\text{-}MC}}^{\pi}  = \inf \left\{y\in\mathbb R:{\mathds P}\left(\hat v_{\mathrm{MB\text{-}MC}}^{\pi,*} - \hat v_{\pi} \le y |{\mathcal{D}} \right) \ge \delta  \right\}.
\end{equation*}
Based on these bootstrap quantiles, we construct a bootstrap confidence interval
\begin{equation*}
\mathrm{CI}_{\mathrm{MB\text{-}MC}}\left( \delta \right) = \left[ \hat v_{\mathrm{MC}}^{\pi}- q_{1-\delta/2, \mathrm{MB\text{-}MC}}^{\pi}, \hat v_{\mathrm{MC}}^{\pi} - q_{\delta/2, \mathrm{MB\text{-}MC}}^{\pi} \right].
\end{equation*}
As a result of Theorem~\ref{Them:PB:MC:asymp_normal}, the coverage probability of the empirical confidence interval of $v_{\pi}$ constructed above converges to the nominal level.
\begin{corollary}[Asymptotic coverage validity] \label{Coro:PB:MC} 
As $n \to \infty$,
\begin{equation*}
{\mathds P}\left(v_{\pi} \in \mathrm{CI}_{\mathrm{MB\text{-}MC}}\left( \delta \right) \right) \to 1 - \delta.
\end{equation*}
\end{corollary}

\subsubsection{Off-policy bootstrap}\label{Sec:3.1.2}

In off-policy inference, the central question is no longer whether to bootstrap, but how to generate bootstrap trajectories under distribution shift. We build on the asymptotic normality of $\hat v_{\mathrm{Plug\text{-}in}}^{\pi}$, and focus on how bootstrap trajectory generation affects the resulting inference procedure. 

Let the offline dataset $\mathcal D$ be collected under a known behavior policy $\mu$, and let $\widehat{\mathcal M}$ be the estimated MDP learned from $\mathcal D$ as shown in (\ref{MDP:hat}). The target remains the value $v_\pi$ of a target policy $\pi$, whose original point estimate is given by $\hat v_{\mathrm{Plug\text{-}in}}^{\pi}$ as shown in (\ref{v_plugin}).

The bootstrap dataset for off-policy is collected by the behavior policy $\mu$, that is 
\begin{equation*}
\widetilde{\mathcal D}_{\mu}^*=\{\xi_{\mu}^{*(j)}\}_{j=1}^n, \quad \xi_{\mu}^{*} \sim (\widehat{\mathcal M}, \mu).
\end{equation*}
Apply the Plug-in estimate with target policy $\pi$ to the regenerated dataset $\widetilde{\mathcal D}_{\mu}^*$, yielding the bootstrap estimator $\hat v_{\mathrm{MB\text{-}(Plug\text{-}in)}}^{\pi,*}$.

\begin{assumption}[Sufficient data coverage]\label{Assump:OPE}
For every $(h,s,a)$ on the target policy support, one has
\begin{equation*}
d_h^\mu(s,a):=\mathbb P_\mu(s_h=s,a_h=a)>0. 
\end{equation*}
\end{assumption}

\begin{theorem}[Distributional consistency] \label{Them:PB:OPE:asymp_normal}
Suppose that Assumption~\ref{Assump:OPE} holds, then the following statement holds:
\begin{equation*}
\sqrt n\Bigl(\hat v_{\mathrm{MB\text{-}(Plug\text{-}in)}}^{\pi,*} - \hat v_{\mathrm{Plug\text{-}in}}^\pi \Bigr) \;\Big|\;\mathcal D \Rightarrow N(0,\sigma_{\mathrm{Plug\text{-}in}}^2).
\end{equation*}
Consequently, it implies
\begin{equation*}
\sup_{t\in\mathbb R}\left|\mathbb P\!\left(\sqrt n\bigl(\hat v_{\mathrm{MB\text{-}(Plug\text{-}in)}}^{\pi,*} - \hat v_{\mathrm{Plug\text{-}in}}^\pi \bigr)\le t \,\middle|\, \mathcal D\right) - \mathbb P\!\left(\sqrt n\bigl(\hat v_{\mathrm{Plug\text{-}in}}^\pi-v_\pi\bigr)\le t \right)\right|\xrightarrow{p}0.
\end{equation*}
\end{theorem}

It follows that the behavior-driven model-based bootstrap yields asymptotically valid confidence intervals and consistent bootstrap variance estimation for $v_\pi$.
\begin{corollary}[Asymptotic coverage validity]\label{Coro:PB:OPE} 
Define the conditional bootstrap quantile
\begin{equation*}
q_{\delta,\mathrm{MB\text{-}(Plug\text{-}in)}}^{\pi} = \inf\left\{y\in\mathbb R:\mathbb P\!\left(\hat v_{\mathrm{MB\text{-}(Plug\text{-}in)}}^{\pi,*} - \hat v_{\mathrm{Plug\text{-}in}}^\pi \le y \mid \mathcal D \right)\ge \delta\right\}.
\end{equation*}
Then the confidence interval
\begin{equation*}
\mathrm{CI}_{\mathrm{MB\text{-}(Plug\text{-}in)}}(\delta) = \left[\hat v_{\mathrm{Plug\text{-}in}}^\pi-q_{1-\delta/2,\mathrm{MB\text{-}(Plug\text{-}in)}}^{\pi},\; \hat v_{\mathrm{Plug\text{-}in}}^\pi-q_{\delta/2,\mathrm{MB\text{-}(Plug\text{-}in)}}^{\pi}\right]
\end{equation*}
satisfies
\begin{equation*}
\mathbb P\!\left(v_\pi\in \mathrm{CI}_{\mathrm{MB\text{-}(Plug\text{-}in)}}(\delta)\right)\to 1-\delta.
\end{equation*}
\end{corollary}

\paragraph{Estimated-behavior-driven bootstrap.}
We now consider the practically important setting in which the offline dataset is generated by a single but unknown behavior policy. In this case, we first estimate the behavior policy from the observed data by $\hat\mu_h(a\mid s)= n_h(s,a)/{n_h(s)}$, and then regenerate a bootstrap dataset
\begin{equation*}
\widetilde{\mathcal D}_{\hat\mu}^* = \{\xi_{\hat\mu}^{*(j)}\}_{j=1}^n, \qquad \xi_{\hat\mu}^{*}\sim (\widehat{\mathcal M},\hat\mu).
\end{equation*}

Here, beyond controlling the discrepancy between the estimated MDP \(\widehat{\mathcal M}\) and \(\mathcal M\), we also need to establish the consistency of the behavior policy estimator: $\hat\mu_h(a|s)\xrightarrow{p}\mu_h(a|s)$. Once the regenerated episode law under \((\widehat{\mathcal M},\hat\mu)\) is shown to consistently approximate its population counterpart under \((\mathcal M,\mu)\), the remaining bootstrap validity argument follows the same line as in the known-behavior case.

\subsection{Offline policy evaluation inference}\label{Sec:3.3}

In this section, we describe how to conduct policy evaluation inference based on the output of above theoretical results. More specifically, the implementation is described in Algorithm~\ref{Alg:MB}.

\begin{itemize}
\item \textbf{Confidence interval}. Compute the $\delta/2$ and $1 - \delta/2$ quantiles of the empirical error distribution
$\left\{\varepsilon_{(1)}, \cdots, \varepsilon_{(B)}\right\}$, denoted as $\hat q_{\delta/2}^{\pi}$, $\hat q_{1 - \delta/2}^{\pi}$, respectively. The bootstrap confidence interval is
\begin{equation}\label{Equ:empirical:CI}
\left[\hat v_{\pi} - \hat q_{1 - \delta/2}^{\pi}, \hat v_{\pi} - \hat q_{\delta/2}^{\pi}\right].
\end{equation}
\item \textbf{Variance estimation}. To estimate the variance of MC and Plug-in estimators, we calculate the sample variance as
\begin{equation}\label{Equ:var:hat}
{\widehat{\mathrm{Var}}}\left(\hat v_{\pi}\right) = \frac{1}{B-1} \sum_{b = 1}^{B}\left(\varepsilon_{(b)} - \bar \varepsilon \right)^2,
\end{equation}
where $\bar \varepsilon = \frac{1}{B}\sum_{b = 1}^{B}\varepsilon_{(b)}$.
\end{itemize}

\begin{algorithm}[h]
\fontsize{10.5}{10}\selectfont
\SetAlgoNoLine
\SetAlgoSkip{0pt}
\SetAlCapSkip{0.3ex}
\caption{Model-based Bootstrap}\label{Alg:MB}
\KwIn{Dataset $\mathcal D=\{\xi^{(i)}=(s_0^{(i)},a_0^{(i)},r_0^{(i)},\ldots,s_H^{(i)})\}_{i=1}^n$, target policy $\pi$, behavior policy $\mu$, confidence level $\delta \in [0,1]$, and bootstrap size $B$.}
\KwOut{$(1-\delta)$ confidence interval for $v_\pi$ and variance estimate $\widehat{\mathrm{Var}}(\hat v_\pi)$.}
Compute the original estimate $\hat v_\pi={\bf F}(\mathcal D)$\tcp*{\footnotesize ${\bf F}$: Plug-in estimate over $\mathcal{D}$}
\For{$b=1,\ldots,B$}{
Generate $\widetilde{\mathcal D}_{(b)}^*=\{\xi_{(b)}^{*(j)}\}_{j=1}^n$ from $(\widehat{\mathcal M},\pi)$ for on-policy, or from $(\widehat{\mathcal M},\mu)$ for off-policy\;
Compute $\hat v_{(b)}^{\pi,*}={\bf U}(\widetilde{\mathcal D}_{(b)}^*)$ \tcp*{\footnotesize ${\bf U}$: MC for on-policy, Plug-in for off-policy}
Set $\varepsilon_{(b)}=\hat v_{(b)}^{\pi,*}-\hat v_\pi$.
}
\Return the bootstrap confidence interval in (\ref{Equ:empirical:CI}) and the variance estimator in (\ref{Equ:var:hat}).
\end{algorithm}

\section{Simulation Studies}\label{Sec:4}

In this section, we conduct numerical experiments to evaluate the performance of the proposed model-based bootstrap (MB) offline policy evaluation method in two representative tabular RL environments: Time-varying MDP (a nonstationary MDP with a fixed horizon) and the Cliff-walking environment (a stationary MDP with a random horizon). The experiments are designed to achieve three main objectives: (1) to compare the accuracy of the error distribution of MC and Plug-in estimate obtained by bootstrapping episodes (BE) and model-based bootstrap (MB) to characterize the true error distributions (see Sec.~\ref{Sec:4.3}); (2) to examine the effectiveness and tightness of the resulting confidence intervals (see Sec.~\ref{Sec:4.4}); (3) to assess the accuracy of the MC and  Plug-in variance estimators (see Sec.~\ref{Sec:4.5}).

\subsection{Simulation environments}\label{Sec:4.1}

\begin{figure}[ht]
\centering
\vspace{-0.35cm}
\tikzset{global scale/.style={scale=#1,every node/.append style={scale=#1}}}
\begin{tikzpicture}[global scale = 0.85]
\draw[-](0,0)--(0,4);
\draw[-](1,0)--(1,4);
\draw[-](2,1)--(2,4);
\draw[-](3,1)--(3,4);
\draw[-](4,1)--(4,4);
\draw[-](5,1)--(5,4);
\draw[-](6,1)--(6,4);
\draw[-](7,1)--(7,4);
\draw[-](8,1)--(8,4);
\draw[-](9,1)--(9,4);
\draw[-](10,1)--(10,4);
\draw[-](11,0)--(11,4);
\draw[-](12,0)--(12,4);

\draw[-](0,0)--(12,0);
\draw[-](0,1)--(12,1);
\draw[-](0,2)--(12,2);
\draw[-](0,3)--(12,3);
\draw[-](0,4)--(12,4);

\draw[fill=gray!50, draw=black](1,0) rectangle (11,1);
\node at (0.5,0.5){\tiny Initial};
\node at (11.5,0.5){\tiny Terminal};
\node at (6,0.5){The Cliff};
\end{tikzpicture}
\caption{Cliff-walking environment.}
\label{Fig:Cliff}
\end{figure}
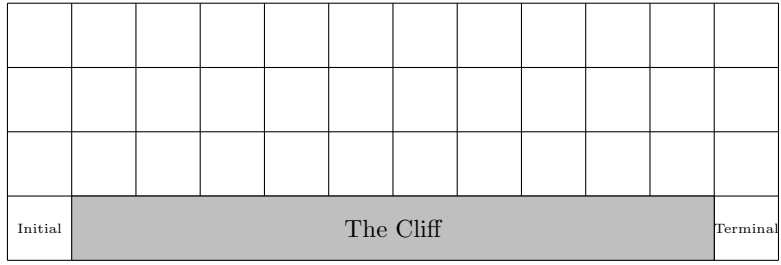

\begin{enumerate}
\item The \textbf{Time-varying MDP} consists of two states $\{s_0,s_1\}$, and two actions $\{a_1,a_2\}$. At each time step, $s_0$ remains unchanged regardless of the chosen action. On the other hand, state $s_1$ transitions to either itself or $s_0$ based on time-varying probabilities. These probabilities are
\begin{equation*}
P_h\left(s_0|s_1,a_1;p_h\right) =\left\{\begin{array}{ll}
2/H, &\hbox{if }p_h < 0.5\\
1, &\hbox{if }p_h \ge 0.5\end{array}\right.
\hbox{ and }P_h\left(s_1|s_1,a_2;p_h\right) = \left\{\begin{array}{ll}1, &\hbox{if }p_h < 0.5\\
1-2/H, &\hbox{if }p_h \ge 0.5
\end{array}\right.,
\end{equation*}

determined by a sequence of i.i.d random numbers $p_h \in U\left[0,1\right],h\in[H]$, as in \cite{Yin:2020} and \cite{Wang:2024}. Here, we modify the immediate reward 1 and 0 to some uniform distributions to make the environment more stochastic. The horizon is set to $H = 10$.

The behavior policy selects both actions with equal probability at each state, while the target policy assigns probability $1/2$ to both actions at $s_0$ and $1/4$ to $a_1$ and $3/4$ to $a_2$ at $s_1$.

\item \textbf{Cliff-walking environment} is a typical tabular MDP with random horizons as shown in Fig.~\ref{Fig:Cliff}. In this environment, the agent begins at the initial state and will be terminated when it reaches the terminal state or falls off the cliff. At each state, four actions are available, namely ${\cal A} = \left\{\mathrm{up,down,left,right}\right\}$. We modify the environment in order to make it more stochastic. Specifically, we introduce randomness in state transitions. Given a state-action pair $\left({s, a}\right)$, the agent moves to the next state indicated by the action with a probability $1-\epsilon$ and the four neighborhoods with an equal probability $\epsilon/4$. Transferring beyond the boundary makes the agent stay where it is. A reward of $-50$ is given for falling off the cliff, and a reward of $-1$ is given for any other transition. We set the parameter $\epsilon$ to $0.4$ to control the stochasticity of the environment.

The target policy is a near-optimal policy trained by Q-learning, whereas the behavior policy is a $0.1$-greedy policy used to generate the offline data.

\end{enumerate}

\subsection{Comparison of error distribution estimation}\label{Sec:4.3}

\begin{figure}[htbp]
\centering
\begin{subfigure}
\centering
\begin{minipage}{\textwidth}
\begin{minipage}{0.1\textwidth}
\centering
(a1)
\end{minipage}
\begin{minipage}{0.28\textwidth}
\centering
\includegraphics[scale=0.24]{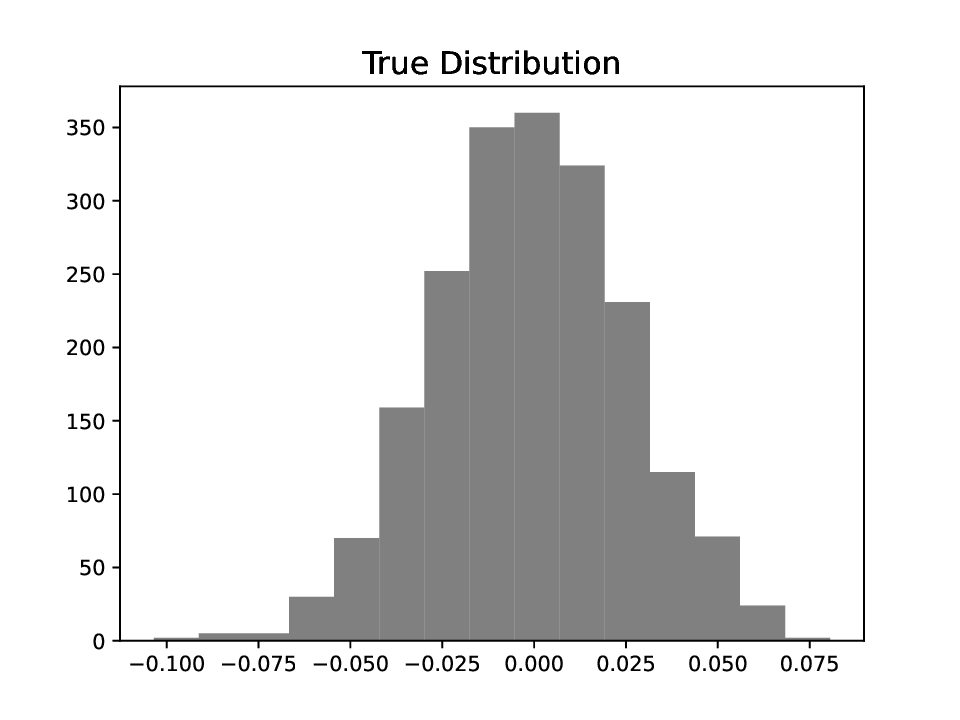}
\end{minipage}
\begin{minipage}{0.28\textwidth}
\centering
\includegraphics[scale=0.24]{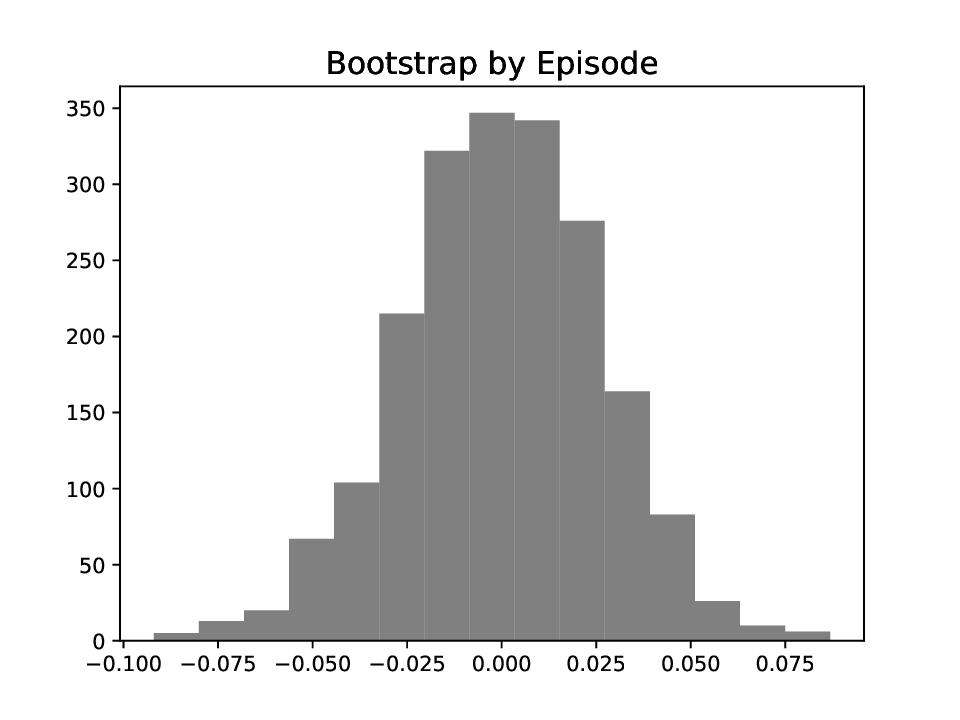}
\end{minipage}
\begin{minipage}{0.28\textwidth}
\centering
\includegraphics[scale=0.24]{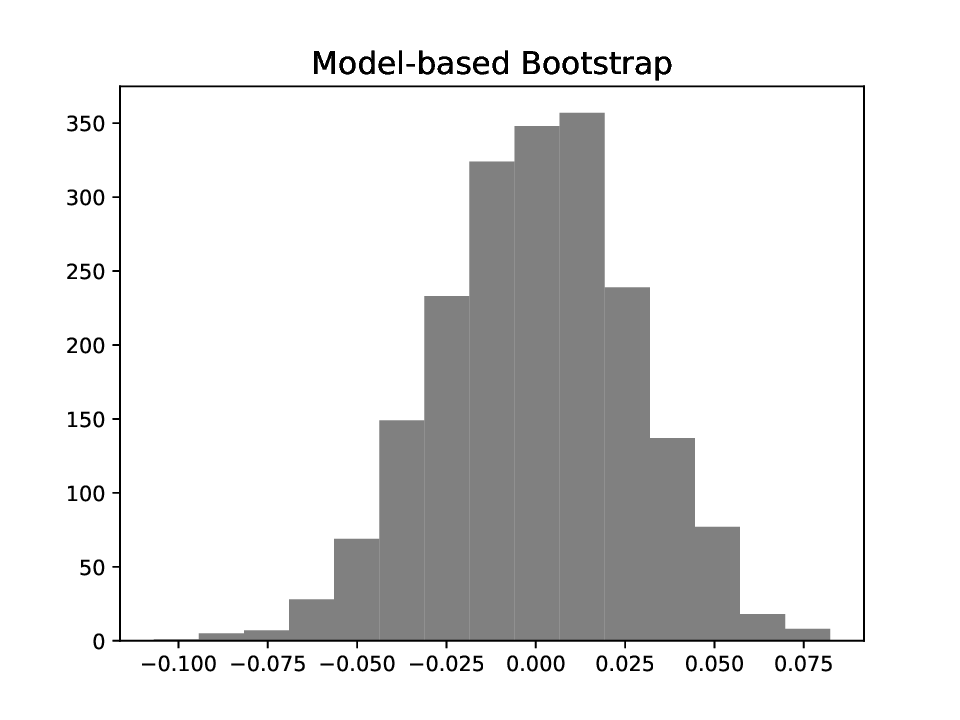}
\end{minipage}
\end{minipage}
\begin{minipage}{\textwidth}
\begin{minipage}{0.1\textwidth}
\centering
(a2)
\end{minipage}
\begin{minipage}{0.28\textwidth}
\centering
\includegraphics[scale=0.24]{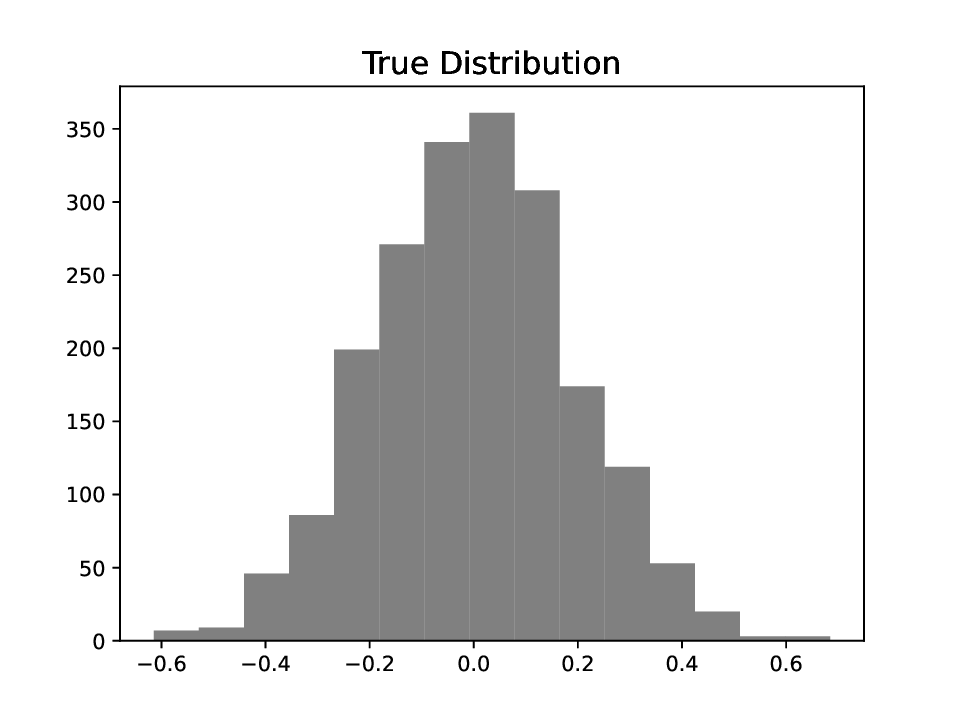}
\end{minipage}
\begin{minipage}{0.28\textwidth}
\centering
\includegraphics[scale=0.24]{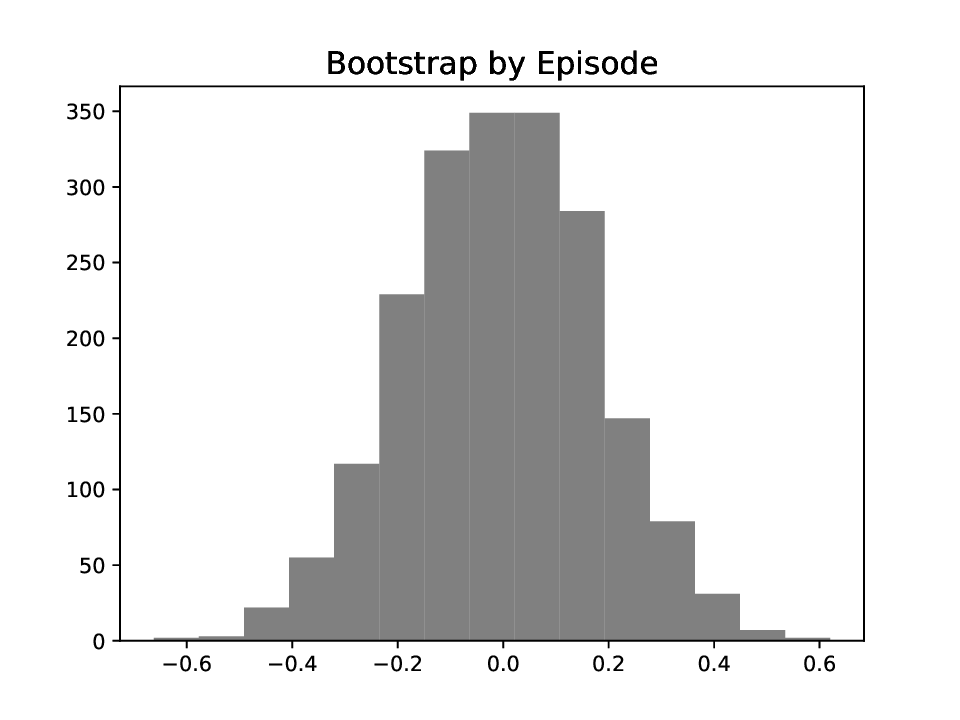}
\end{minipage}
\begin{minipage}{0.28\textwidth}
\centering
\includegraphics[scale=0.24]{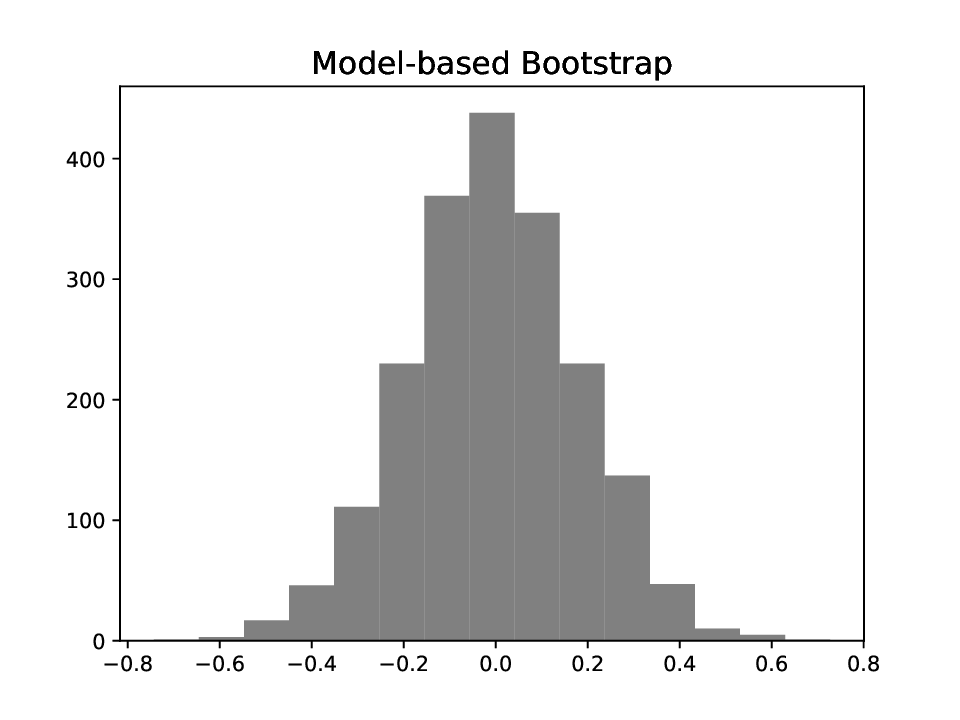}
\end{minipage}
\end{minipage}
\par\vspace{1mm}
{\centering (a) MC estimation error distributions\par}
\vspace{1mm}
\end{subfigure}

\begin{subfigure}
\centering
\begin{minipage}{\textwidth}
\begin{minipage}{0.1\textwidth}
\centering
(b1)
\end{minipage}
\begin{minipage}{0.28\textwidth}
\centering
\includegraphics[scale=0.24]{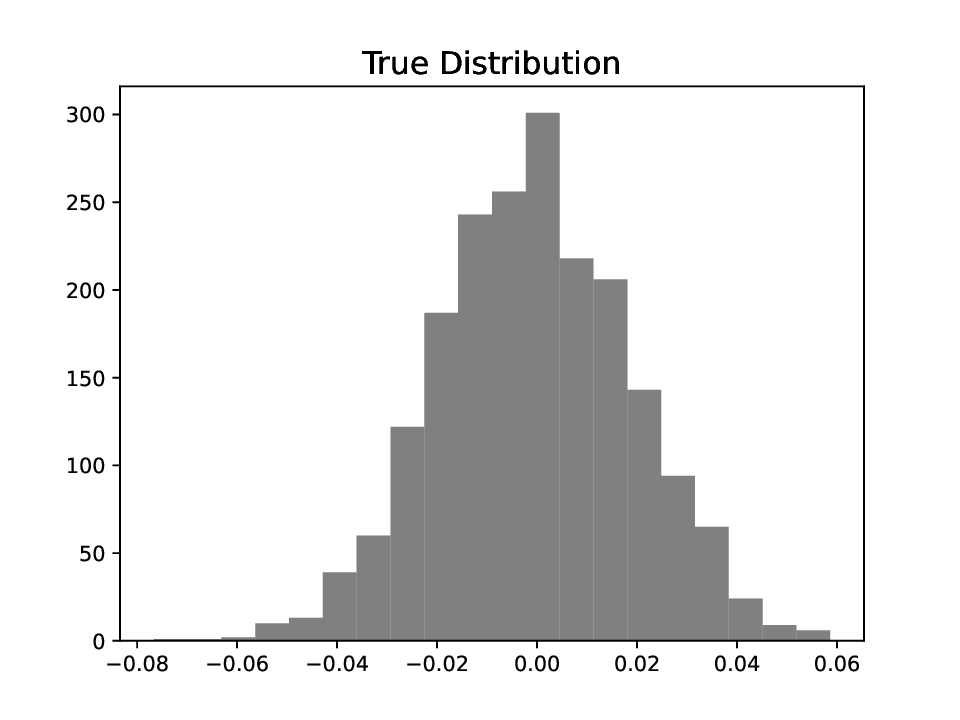}
\end{minipage}
\begin{minipage}{0.28\textwidth}
\centering
\includegraphics[scale=0.24]{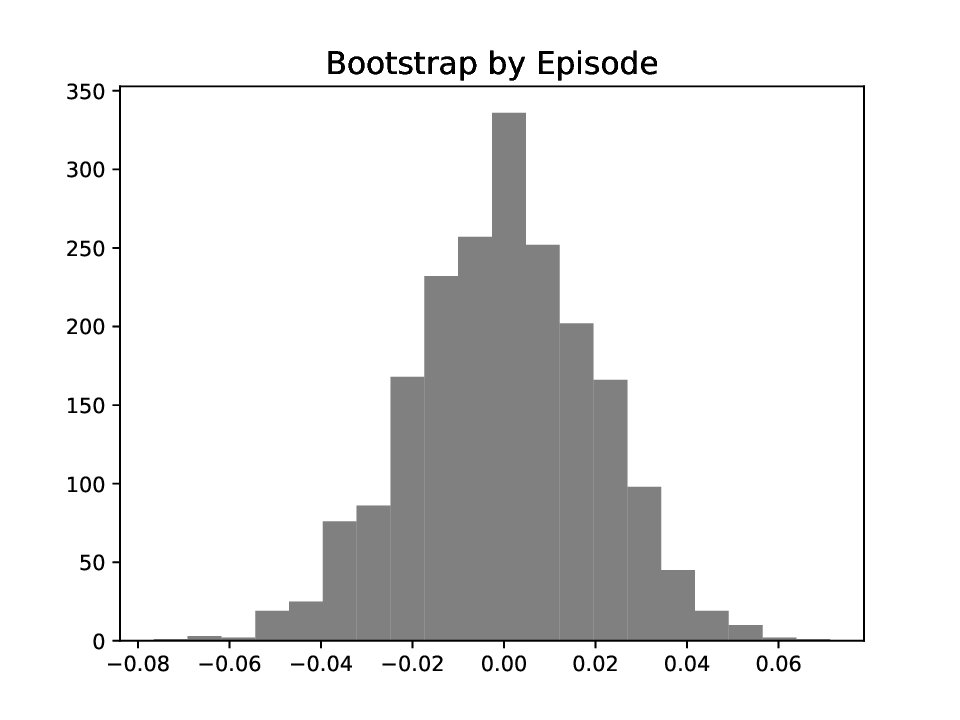} 
\end{minipage}
\begin{minipage}{0.28\textwidth}
\centering
\includegraphics[scale=0.24]{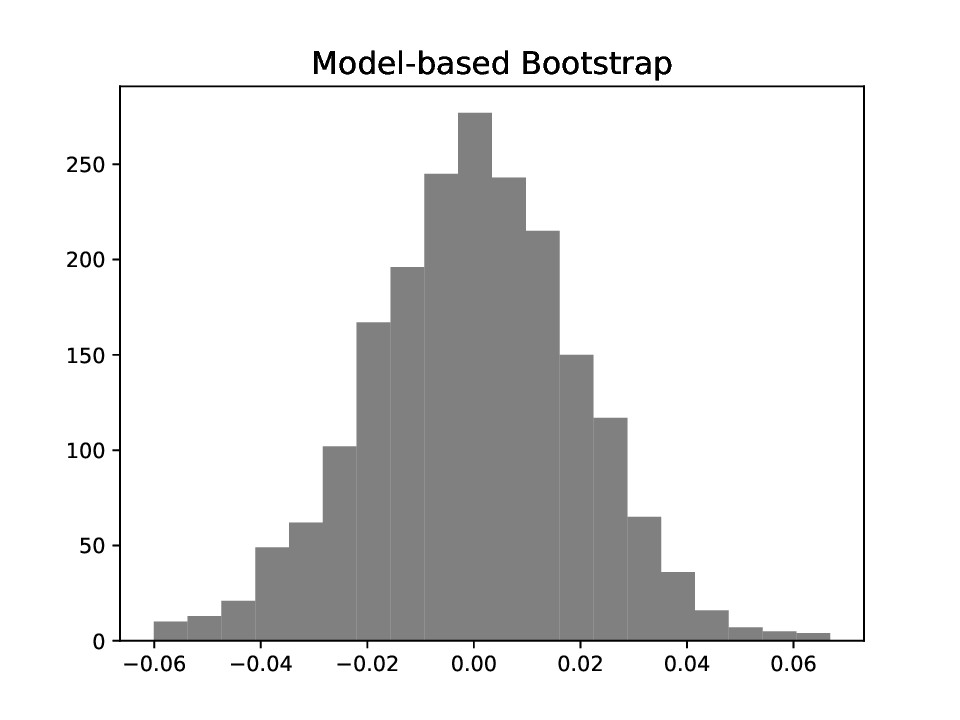}
\end{minipage}
\end{minipage}
\begin{minipage}{\textwidth}
\begin{minipage}{0.1\textwidth}
\centering
(b2)
\end{minipage}
\begin{minipage}{0.28\textwidth}
\centering
\includegraphics[scale=0.24]{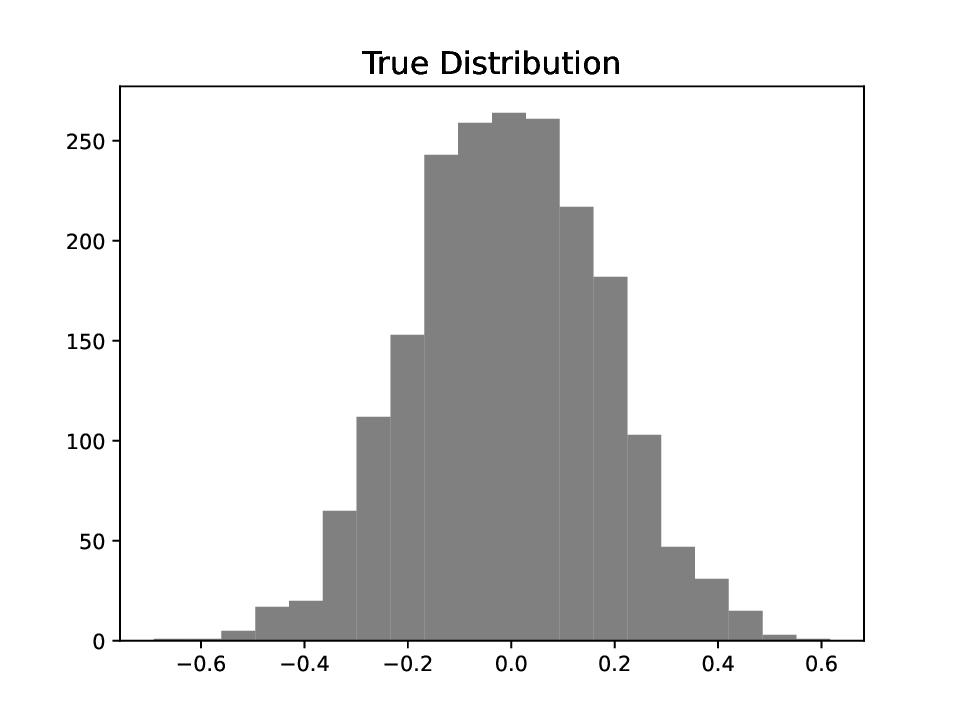}
\end{minipage}
\begin{minipage}{0.28\textwidth}
\centering
\includegraphics[scale=0.24]{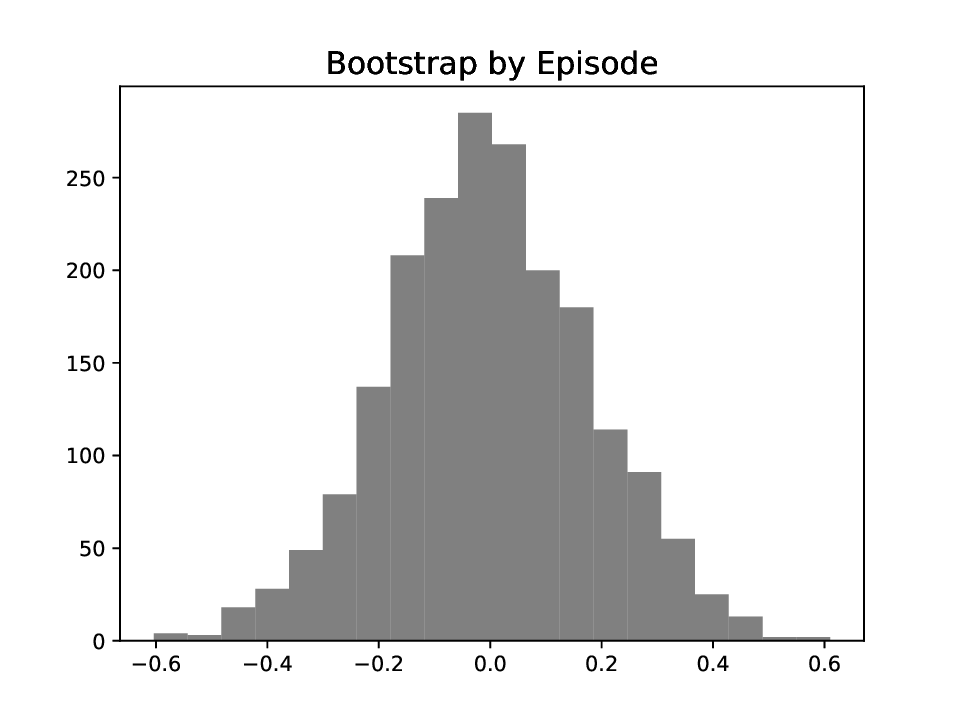}
\end{minipage}
\begin{minipage}{0.28\textwidth}
\centering
\includegraphics[scale=0.24]{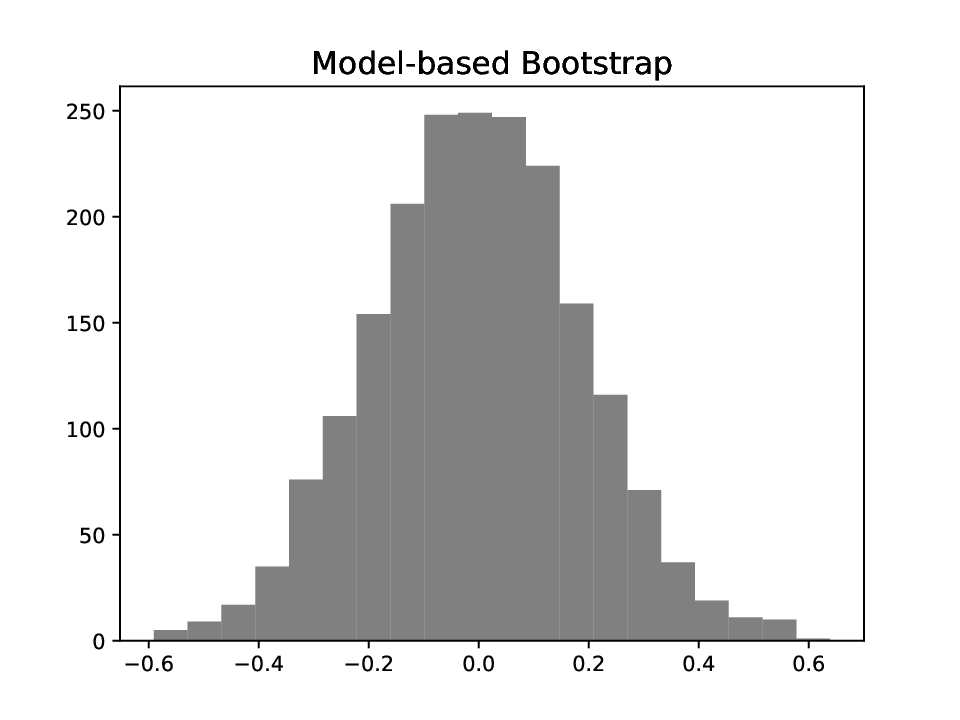}
\end{minipage}
\end{minipage}
\par\vspace{1mm}
{\centering (b) Plug-in estimation error distributions\par}
\vspace{1mm}
\end{subfigure}


\begin{subfigure}
\centering
\begin{minipage}{\textwidth}
\begin{minipage}{0.1\textwidth}
\centering 
(c1)
\end{minipage}
\begin{minipage}{0.28\textwidth}
\centering
\includegraphics[scale=0.24]{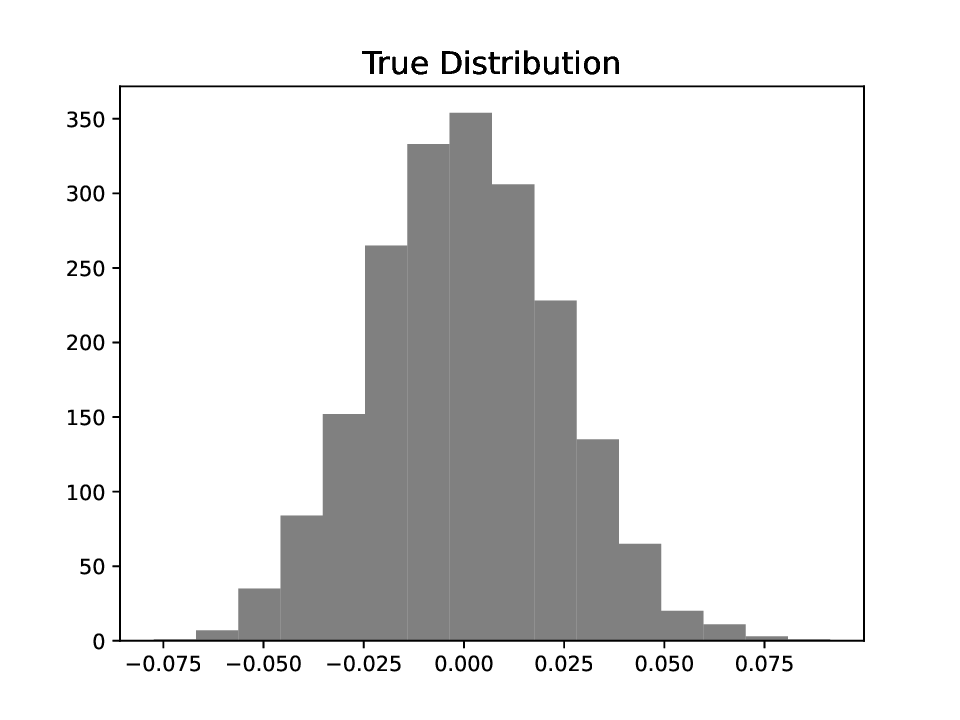}
\end{minipage}
\begin{minipage}{0.28\textwidth}
\centering
\includegraphics[scale=0.24]{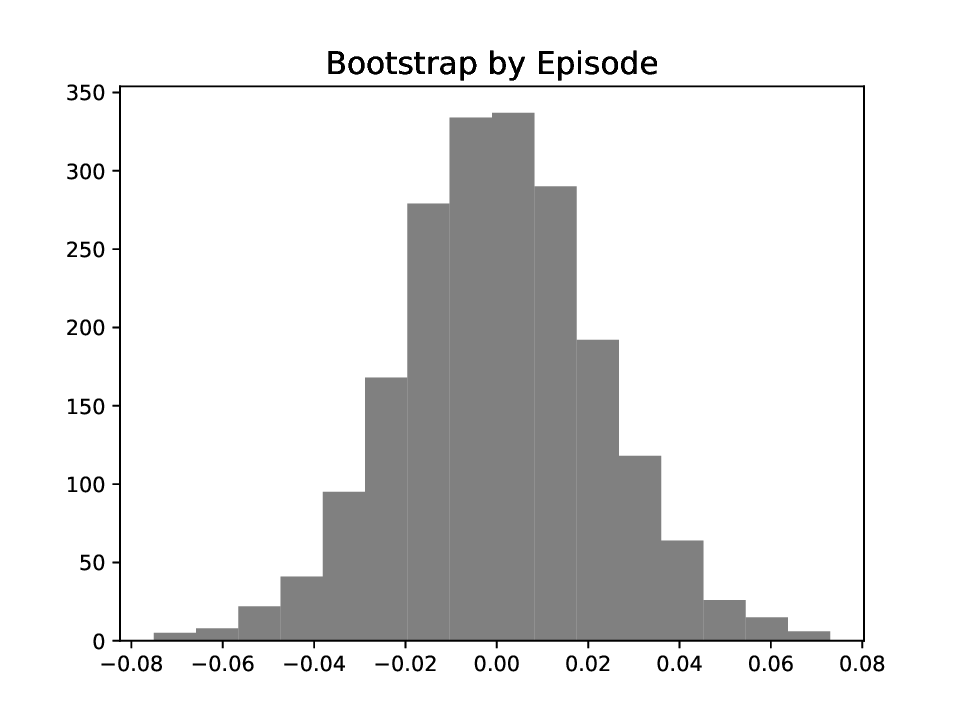}
\end{minipage}
\begin{minipage}{0.28\textwidth}
\centering
\includegraphics[scale=0.24]{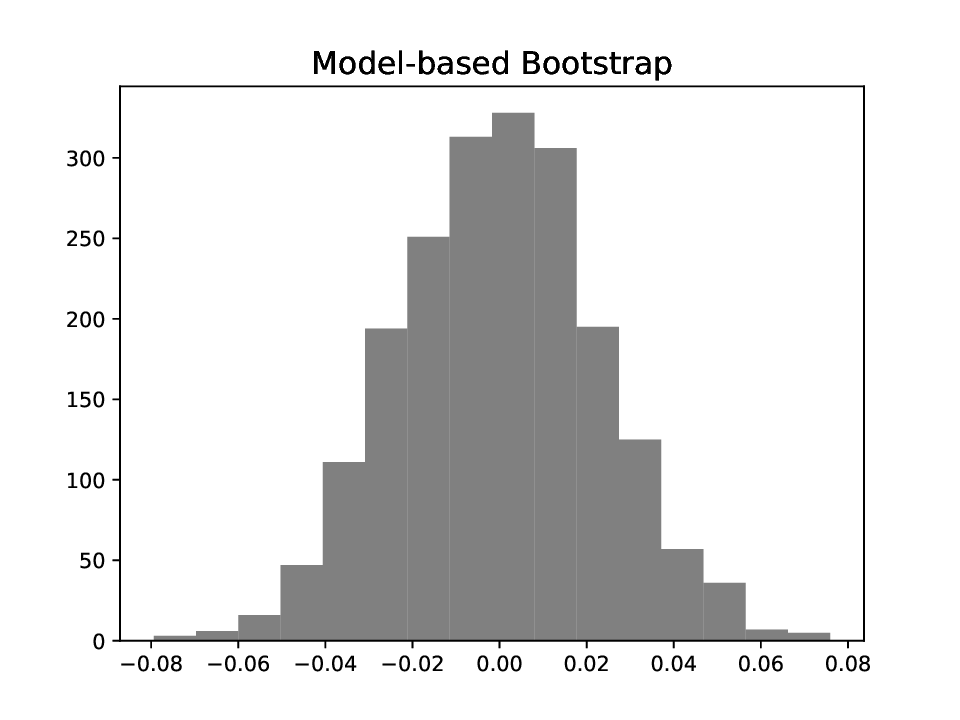}
\end{minipage}
\end{minipage}
\begin{minipage}{\textwidth}
\begin{minipage}{0.1\textwidth}
\centering
(c2)
\end{minipage}
\begin{minipage}{0.28\textwidth}
\centering
\includegraphics[scale=0.24]{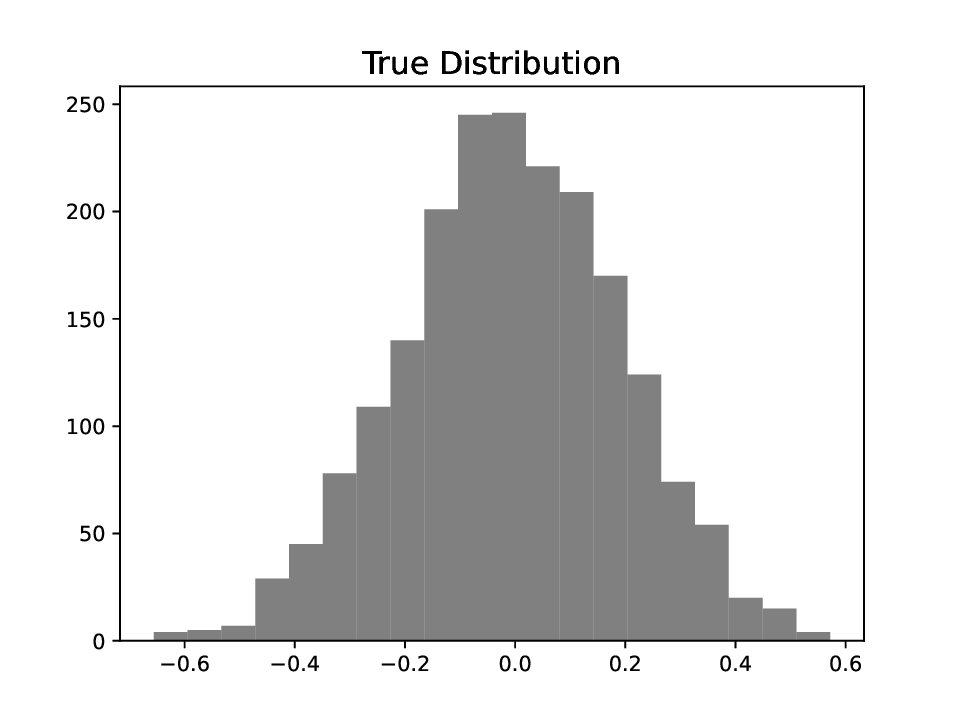}
\end{minipage}
\begin{minipage}{0.28\textwidth}
\centering
\includegraphics[scale=0.24]{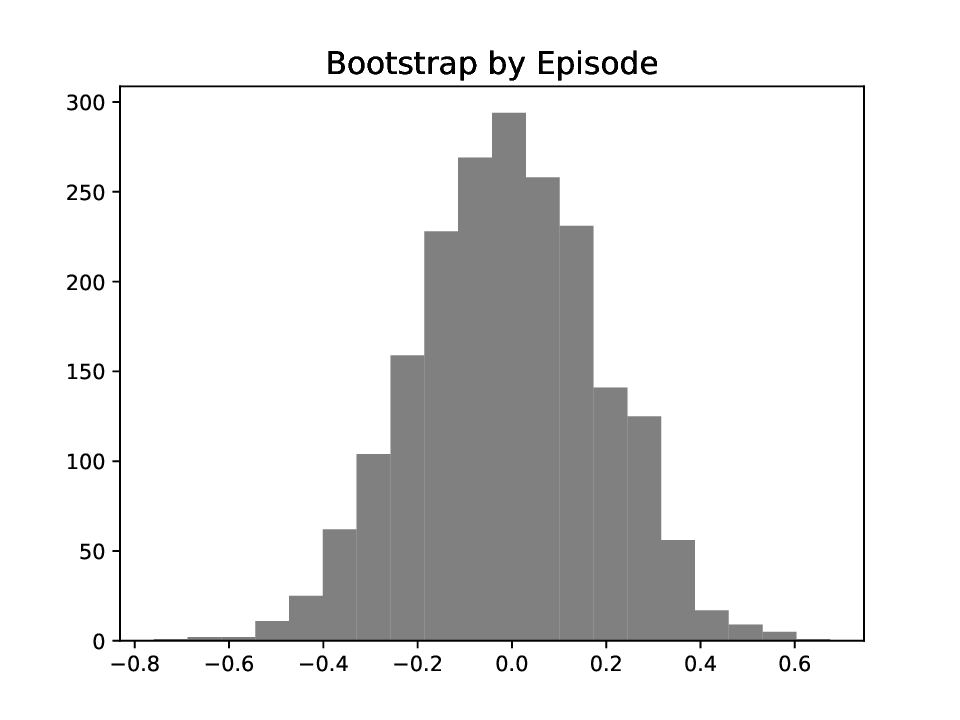}
\end{minipage}
\begin{minipage}{0.28\textwidth}
\centering
\includegraphics[scale=0.24]{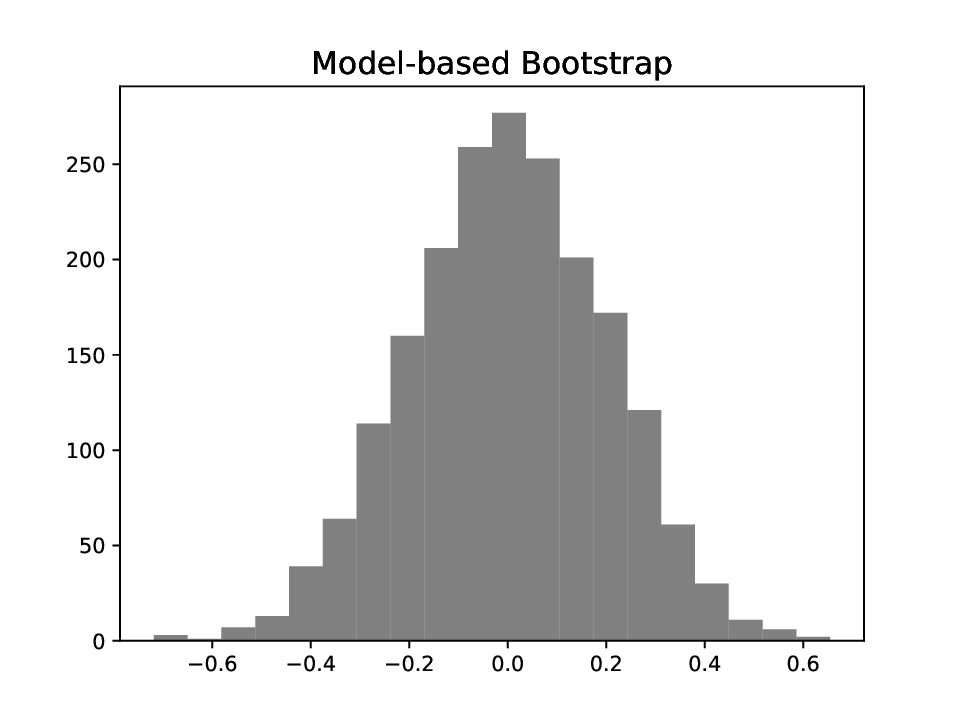}
\end{minipage}
\end{minipage}
\par\vspace{1mm}
{\centering (c) Off-policy Plug-in estimation error distributions\par}
\vspace{1mm}
\end{subfigure}
\caption{Simulation results of MC and Plug-in estimation error distributions in the Time-varying MDP and Cliff-walking environment.}
\label{Fig:Error:distribution}
\end{figure}

This section investigates how the two resampling ways -- BE and MB -- approximate the true error distributions of MC and Plug-in policy evaluators, respectively. The ground-truth error distributions are obtained via Monte Carlo simulation with a sample size of 10,000, and the number of bootstrap replications is set to 2,000. Fig.~\ref{Fig:Error:distribution} (a) and (b) present the MC and Plug-in on-policy estimate error distributions, respectively, and Fig.~\ref{Fig:Error:distribution} (c) reports the off-policy Plug-in error distributions. The results demonstrate that both BE- and MB-based bootstrap distributions closely match the true distribution of the evaluation error $\hat v_{\pi} - v_{\pi}$.

\subsection{Comparison of confidence intervals}\label{Sec:4.4}

In this section, we evaluate the performance of MB in confidence interval construction in the above two RL environments. We study the empirical coverage probability and interval width with different sample sizes. The confidence intervals are constructed at significance levels $\delta \in \{0.25, 0.1, 0.05\}$. The number of bootstrap resampling and replications are all set to be 100. The simulation results shown in Tab.~\ref{tab:timevary_results}--\ref{tab:cliff_offpolicy} lead to the following observations:
\begin{itemize}
\item[(1)] Tab.~\ref{tab:timevary_results} and~\ref{tab:timevary_fqe_results} report the empirical coverage probabilities and average interval widths in the Time-varying environment. Across both on-policy and off-policy settings, all methods exhibit decreasing interval widths as the sample size increases, confirming that the uncertainty of policy value estimation diminishes with more trajectories. 

In the on-policy setting, MB-(Plug-in) performs comparably to BE-(Plug-in) and BT-(Plug-in), while often producing shorter intervals with empirical coverage close to the nominal level, especially at the practically relevant $0.90$ and $0.95$ confidence levels. The MC-based methods, BE-MC and MB-MC, show similar but slightly inferior behavior, as MC is not asymptotically optimal. 

In the off-policy setting, the advantage of MB-(Plug-in) is more pronounced. Although distribution shift makes inference more challenging and generally leads to wider intervals, MB-(Plug-in) yields substantially shorter intervals than BE-(Plug-in) and BT-(Plug-in), while maintaining acceptable coverage accuracy, especially when the sample size is small or moderate. For instance, at the $0.95$ level with $n=100$, the average interval width of MB-(Plug-in) is $1.3641$, compared with $1.5261$ for BE-(Plug-in) and $1.7652$ for BT-(Plug-in). This indicates that MB-FQE achieves higher finite-sample statistical efficiency.

\item[(2)] Tab.~\ref{tab:cliff_onpolicy} and~\ref{tab:cliff_offpolicy} summarize the results for the Cliff-walking environment. In the on-policy setting, all methods show decreasing interval widths as the sample size increases. MB-(Plug-in) achieves coverage close to the nominal levels while producing competitive interval widths, especially at the $0.90$ and $0.95$ levels. 

In the off-policy setting, distribution shift leads to substantially wider confidence intervals, particularly for small sample sizes. BE-(Plug-in) and BT-(Plug-in) are more conservative, with BT-(Plug-in) producing especially wide intervals when $n$ is small. MB-(Plug-in) yields much shorter intervals and thus shows a clear efficiency advantage, although it may suffer from finite-sample under-coverage under severe distribution shift. Its coverage improves as the sample size increases and approaches the nominal level.
\end{itemize}

All in all, the extensive numerical simulation studies confirm the effectiveness and advantages of MB method in offline policy interval estimation.

\begin{table}[htbp]
\centering
\captionsetup[table]{skip=0.6pt}
\caption{Time-varying (on-policy): empirical coverage probability and average interval width. Each entry is reported as Coverage / Width.}
\renewcommand{\arraystretch}{0.6}
\label{tab:timevary_results}
\resizebox{\textwidth}{!}{
\begin{tabular}{llccccc}
\toprule
Method & Level & $n=50$ & $n=100$ & $n=200$ & $n=500$ & $n=1000$ \\
\midrule
& $0.75$ & 0.78 / 0.8561 & 0.76 / 0.6235 & 0.81 / 0.4338 & 0.77 / 0.2753 & 0.71 / 0.1979 \\
BE-MC & $0.90$ & 0.91 / 1.2166 & 0.91 / 0.8972 & 0.93 / 0.6207 & 0.89 / 0.3927 & 0.90 / 0.2814 \\
& $0.95$ & 0.95 / 1.4495 & 0.96 / 1.0634 & 0.97 / 0.7395 & 0.95 / 0.4701 & 0.94 / 0.3329 \\
\midrule
& $0.75$ & 0.78 / 0.8101 & 0.74 / 0.4765 & 0.69 / 0.3280 & 0.74 / 0.2054 & 0.69 / 0.1454 \\
BE-(Plug-in) & $0.90$ & 0.89 / 1.2208 & 0.88 / 0.6874 & 0.89 / 0.4695 & 0.86 / 0.2943 & 0.87 / 0.2069 \\
& $0.95$ & 0.93 / 1.4935 & 0.95 / 0.8230 & 0.93 / 0.5588 & 0.91 / 0.3507 & 0.95 / 0.2496 \\
\midrule
& $0.75$ & 0.81 / 0.9035 & 0.74 / 0.4845 & 0.70 / 0.3302 & 0.73 / 0.2066 & 0.64 / 0.1460 \\
BT-(Plug-in) & $0.90$ & 0.90 / 1.3676 & 0.88 / 0.6987 & 0.89 / 0.4711 & 0.88 / 0.2960 & 0.86 / 0.2089 \\
& $0.95$ & 0.96 / 1.6827 & 0.98 / 0.8480 & 0.94 / 0.5584 & 0.94 / 0.3534 & 0.96 / 0.2497 \\
\midrule
& $0.75$ & 0.66 / 0.6734 & 0.74 / 0.4600 & 0.71 / 0.3241 & 0.72 / 0.2063 & 0.65 / 0.1456 \\
MB-(Plug-in) & $0.90$ & 0.87 / 0.9777 & 0.85 / 0.6598 & 0.87 / 0.4694 & 0.88 / 0.2946 & 0.86 / 0.2070 \\
& $0.95$ & 0.96 / 1.1837 & 0.95 / 0.7868 & 0.94 / 0.5622 & 0.92 / 0.3508 & 0.95 / 0.2462 \\
\midrule
& $0.75$ & 0.79 / 0.8689 & 0.72 / 0.6137 & 0.83 / 0.4341 & 0.76 / 0.2791 & 0.74 / 0.1948 \\
MB-MC & $0.90$ & 0.92 / 1.2443 & 0.92 / 0.8826 & 0.91 / 0.6219 & 0.89 / 0.3975 & 0.90 / 0.2778 \\
& $0.95$ & 0.94 / 1.4804 & 0.95 / 1.0545 & 0.98 / 0.7412 & 0.94 / 0.4725 & 0.94 / 0.3303 \\
\bottomrule
\end{tabular}
}
\end{table}
\begin{table}[htbp]
\centering
\captionsetup[table]{skip=0.6pt}
\caption{Time-varying (off-policy): empirical coverage probability and average interval width. Each entry is reported as Coverage / Width.}
\renewcommand{\arraystretch}{0.6}
\label{tab:timevary_fqe_results}
\resizebox{\textwidth}{!}{
\begin{tabular}{llccccc}
\toprule
Method & Level & $n=50$ & $n=100$ & $n=200$ & $n=500$ & $n=1000$ \\
\midrule
& $0.75$ & 0.79 / 1.5465 & 0.76 / 0.8706 & 0.82 / 0.5587 & 0.76 / 0.3214 & 0.77 / 0.2233 \\
BE-(Plug-in) & $0.90$ & 0.88 / 2.3545 & 0.89 / 1.2610 & 0.89 / 0.8203 & 0.87 / 0.4612 & 0.91 / 0.3234 \\
& $0.95$ & 0.91 / 2.9800 & 0.92 / 1.5261 & 0.95 / 0.9950 & 0.96 / 0.5556 & 0.95 / 0.3884 \\
\midrule
& $0.75$ & 0.81 / 1.7004 & 0.79 / 0.9427 & 0.77 / 0.5826 & 0.74 / 0.3214 & 0.78 / 0.2267 \\
BT-(Plug-in) & $0.90$ & 0.87 / 2.5893 & 0.91 / 1.4160 & 0.93 / 0.8510 & 0.91 / 0.4629 & 0.91 / 0.3271 \\
& $0.95$ & 0.93 / 3.2895 & 0.93 / 1.7652 & 0.95 / 1.0623 & 0.97 / 0.5582 & 0.94 / 0.3884 \\
\midrule
& $0.75$ & 0.77 / 1.1785 & 0.72 / 0.7737 & 0.75 / 0.5217 & 0.75 / 0.3232 & 0.75 / 0.2292 \\
MB-(Plug-in) & $0.90$ & 0.89 / 1.7392 & 0.86 / 1.1235 & 0.88 / 0.7479 & 0.92 / 0.4592 & 0.90 / 0.3288 \\
& $0.95$ & 0.89 / 2.1339 & 0.93 / 1.3641 & 0.94 / 0.8956 & 0.92 / 0.5478 & 0.95 / 0.3944 \\
\bottomrule
\end{tabular}
}
\end{table}

\begin{table}[htbp]
\centering
\captionsetup[table]{skip=0.6pt}
\caption{Cliff-walking (on-policy): empirical coverage probability and average interval width. Each entry is reported as Coverage / Width.}
\renewcommand{\arraystretch}{0.6}
\label{tab:cliff_onpolicy}
\resizebox{\textwidth}{!}{
\begin{tabular}{llccccc}
\toprule
Method & Level & $n=10$ & $n=50$ & $n=100$ & $n=200$ & $n=500$ \\
\midrule
& $0.75$ & 0.66 / 12.2985 & 0.72 / 5.9936 & 0.73 / 4.1815 & 0.75 / 3.0496 & 0.75 / 1.8777 \\
BE-MC & $0.90$ & 0.79 / 17.4210 & 0.90 / 8.4850 & 0.89 / 6.0345 & 0.88 / 4.3553 & 0.89 / 2.7034 \\
& $0.95$ & 0.88 / 20.8415 & 0.93 / 10.1985 & 0.93 / 7.2514 & 0.94 / 5.1888 & 0.93 / 3.2301 \\
\midrule
& $0.75$ & 0.64 / 11.9818 & 0.68 / 5.8675 & 0.71 / 4.2237 & 0.73 / 2.9391 & 0.80 / 1.8997 \\
BE-(Plug-in) & $0.90$ & 0.81 / 17.0041 & 0.87 / 8.3158 & 0.88 / 6.0918 & 0.88 / 4.2192 & 0.89 / 2.7557 \\
& $0.95$ & 0.88 / 20.0601 & 0.95 / 10.0290 & 0.94 / 7.2759 & 0.93 / 5.0268 & 0.95 / 3.2815\\
\midrule
& $0.75$ & 0.93 / 36.9530 & 0.80 / 6.9459 & 0.76 / 4.4046 & 0.73 / 2.9856 & 0.76 / 1.8883 \\
BT-(Plug-in) & $0.90$ & 0.99 / 79.1637 & 0.94 / 12.9712 & 0.91 / 6.5181 & 0.87 / 4.3052 & 0.90 / 2.6942 \\
& $0.95$ & 1.00 / 117.3243 & 0.97 / 26.6507 & 0.96 / 8.4290 & 0.95 / 5.2033 & 0.94 / 3.2238 \\
\midrule
& $0.75$ & 0.64 / 12.4012 & 0.73 / 5.8929 & 0.76 / 4.1429 & 0.72 / 2.9835 & 0.76 / 1.9303 \\
MB-(Plug-in) & $0.90$ & 0.78 / 17.5708 & 0.89 / 8.4319 & 0.88 / 5.9971 & 0.90 / 4.2695 & 0.92 / 2.7220 \\
& $0.95$ & 0.89 / 20.8501 & 0.94 / 10.0468 & 0.95 / 7.2251 & 0.95 / 5.0948 & 0.95 / 3.2301 \\
\midrule
& $0.75$ & 0.66 / 12.8410 & 0.70 / 5.9955 & 0.72 / 4.2637 & 0.76 / 2.9879 & 0.76 / 1.8801 \\
MB-MC & $0.90$ & 0.87 / 17.9860 & 0.90 / 8.5994 & 0.90 / 6.1361 & 0.91 / 4.3061 & 0.89 / 2.7084 \\
& $0.95$ & 0.91 / 21.6290 & 0.93 / 10.3744 & 0.93 / 7.3029 & 0.96 / 5.1521 & 0.96 / 3.2329 \\
\bottomrule
\end{tabular}
}
\end{table}
\begin{table}[htbp]
\centering
\captionsetup[table]{skip=0.6pt}
\caption{Cliff-walking (off-policy): empirical coverage probability and average interval width. Each entry is reported as Coverage / Width.}
\renewcommand{\arraystretch}{0.6}
\label{tab:cliff_offpolicy}
\resizebox{\textwidth}{!}{
\begin{tabular}{llccccc}
\toprule
Method & Level & $n=50$ & $n=100$ & $n=200$ & $n=500$ & $n=1000$ \\
\midrule
& $0.75$ & 0.72 / 178.5161 & 0.84 / 11.5095 & 0.63 / 3.2889 & 0.74 / 2.0457 & 0.75 / 1.4331 \\
BE-(Plug-in) & $0.90$ & 0.90 / 470.0272 & 0.97 / 37.3836 & 0.86 / 4.6406 & 0.89 / 2.9168 & 0.90 / 2.0659 \\
& $0.95$ & 0.93 / 943.9822 & 0.99 / 66.5477 & 0.94 / 5.5635 & 0.93 / 3.5171 & 0.94 / 2.4708 \\
\midrule
& $0.75$ & 0.78 / 331.1729 & 0.85 / 13.8215 & 0.68 / 3.2364 & 0.75 / 2.0699 & 0.81 / 1.4709 \\
BT-(Plug-in) & $0.90$ & 0.92 / 986.3986 & 0.94 / 68.6383 & 0.87 / 4.6177 & 0.91 / 2.9364 & 0.90 / 2.0879 \\
& $0.95$ & 0.99 / 2370.0668 & 1.00 / 263.9416 & 0.90 / 5.5442 & 0.95 / 3.4890 & 0.95 / 2.5196 \\
\midrule
& $0.75$ & 0.37 / 73.6656 & 0.64 / 4.3549 & 0.63 / 3.1060 & 0.73 / 2.0727 & 0.77 / 1.4524 \\
MB-(Plug-in) & $0.90$ & 0.58 / 279.7528 & 0.82 / 12.0482 & 0.81 / 4.4200 & 0.89 / 2.9721 & 0.88 / 2.0848 \\
& $0.95$ & 0.72 / 687.4008 & 0.86 / 95.9276 & 0.88 / 5.5729 & 0.95 / 3.5011 & 0.94 / 2.5026 \\
\bottomrule
\end{tabular}
}
\end{table}

\subsection{Comparison of variance estimation}\label{Sec:4.5}

In this section, we investigate the performance of variance estimation for the MC and Plug-in policy evaluators in both on-policy and off-policy settings. The ground-truth variance, \(\mathrm{Var}(\hat v_{\pi})\), is approximated via a large-scale Monte Carlo procedure. For each method, we evaluate the variance estimation error $\left|\widehat{\mathrm{Var}}(\hat v_{\pi}) - \mathrm{Var}(\hat v_{\pi})\right|$ using \(B=300\) bootstrap resamples over 100 independent replications. The simulation results are reported in Tab.~\ref{tab:var_1}--\ref{tab:var_3}. In these tables, we summarize the errors in the form of median \([Q_1,Q_3]\), where \(Q_1\) and \(Q_3\) denote the first and third quartiles, respectively. For completeness, the corresponding mean (SD) results are provided in the Appendix. The simulation results reveal several clear patterns:
\begin{itemize}
\item[(1)] First, across all settings, the variance estimation error generally decreases as the sample size increases, which is consistent with the asymptotic variance consistency. 
\item[(2)] For on-policy MC estimation (Tab.~\ref{tab:var_1}), MB-MC consistently outperforms or matches BE-MC in both environments, indicating that the model-based bootstrap approach yields more accurate variance estimates. For on-policy Plug-in estimation (Tab.~\ref{tab:var_2}), MB-(Plug-in) achieves the best overall performance, while BE-(Plug-in) is generally competitive. By contrast, BT-(Plug-in) is highly unstable in the Cliff-walking environment when the sample size is small, although its performance improves as $n$ grows. In the Time-varying MDP, the differences among the three methods become much smaller, but MB-(Plug-in) remains the most accurate or nearly the most accurate throughout.
\item[(3)] For off-policy Plug-in estimation (Tab.~\ref{tab:var_3}), the contrast between the two environments is even more pronounced.  In the Time-varying MDP, MB-(Plug-in) attains the smallest error across nearly all sample sizes, demonstrating clear robustness and accuracy in the off-policy setting. In the Cliff-walking environment, all methods incur substantial variance estimation errors in small samples, with BT-(Plug-in) being particularly unstable and exhibiting extremely large errors. Although BE-(Plug-in) and MB-(Plug-in) are also inaccurate when $n$ is small, they remain far more stable than BT-(Plug-in).
\end{itemize}
Overall, the results suggest that the MB-based method consistently outperforms the alternatives in most scenarios, with particularly pronounced advantages in the small-sample regime.

\begin{table}[htbp]
\centering
\captionsetup[table]{skip=0.6pt}
\caption{Variance estimation error of on-policy MC estimate under different sample sizes. Entries are reported as median [Q1, Q3].}
\renewcommand{\arraystretch}{0.5}
\label{tab:var_1}
\begin{tabular}{lcc}
\toprule
Sample size $n$ & BE-MC & MB-MC \\
\midrule
\multicolumn{3}{c}{Panel A: Time-varying MDP} \\
\midrule
50   & 0.022 [0.010, 0.032] & 0.011 [0.005, 0.019] \\
100  & 0.010 [0.004, 0.016] & 0.005 [0.002, 0.008] \\
200  & 0.004 [0.002, 0.006] & 0.003 [0.001, 0.004] \\
500  & 0.001 [0.001, 0.002] & 0.001 [0.000, 0.001] \\
1000 & 0.001 [0.000, 0.001] & 0.000 [0.000, 0.001] \\
\midrule
\multicolumn{3}{c}{Panel B: Cliff-walking environment} \\
\midrule
10   & 8.392 [4.745, 13.168] & 6.841 [2.977, 10.833] \\
50   & 0.807 [0.375, 1.274] & 0.652 [0.278, 1.058] \\
100  & 0.282 [0.135, 0.487] & 0.236 [0.128, 0.375] \\
200  & 0.122 [0.060, 0.224] & 0.093 [0.044, 0.166] \\
500  & 0.048 [0.023, 0.071] & 0.040 [0.023, 0.069] \\
\bottomrule
\end{tabular}
\end{table}

\begin{table}[htbp]
\centering
\captionsetup[table]{skip=0.6pt}
\caption{Variance estimation error of on-policy Plug-in estimate under different sample sizes. Entries are reported as median [Q1, Q3].}
\renewcommand{\arraystretch}{0.5}
\label{tab:var_2}
\setlength{\tabcolsep}{3pt}
\begin{tabular}{lccc}
\toprule
Sample size $n$ & BE-(Plug-in) & BT-(Plug-in) & MB-(Plug-in) \\
\midrule
\multicolumn{4}{c}{Panel A: Time-varying MDP} \\
\midrule
50   & 0.046 [0.021, 0.081] & 0.070 [0.043, 0.118] & 0.011 [0.005, 0.016] \\
100  & 0.005 [0.002, 0.008]  & 0.005 [0.003, 0.009]  & 0.003 [0.001, 0.005]  \\
200  & 0.002 [0.001, 0.002]  &  0.002 [0.001, 0.002] &  0.001 [0.000, 0.002] \\
500  & 0.001 [0.000, 0.001]  &  0.000 [0.000, 0.001] & 0.000 [0.000, 0.001]  \\
1000 & 0.000 [0.000, 0.000] &  0.000 [0.000, 0.000] & 0.000 [0.000, 0.000] \\
\midrule
\multicolumn{4}{c}{Panel B: Cliff-walking environment} \\
\midrule
10   & 7.127 [2.972, 11.298] & 853.455 [505.645, 1294.019] & 5.739 [2.645, 10.286] \\
50   & 0.761 [0.381, 1.279]  & 30.153 [5.655, 117.543]     & 0.550 [0.217, 0.936] \\
100  & 0.257 [0.160, 0.419]  & 0.451 [0.194, 1.390]        & 0.246 [0.119, 0.384] \\
200  & 0.144 [0.067, 0.217]  & 0.123 [0.056, 0.209]        & 0.117 [0.049, 0.204] \\
500  & 0.047 [0.022, 0.072]  & 0.041 [0.025, 0.076]        & 0.035 [0.018, 0.067] \\
\bottomrule
\end{tabular}
\end{table}

\begin{table}[htbp]
\centering
\captionsetup[table]{skip=0.6pt}
\caption{Variance estimation error of off-policy Plug-in estimate under different sample sizes. Entries are reported as median [Q1, Q3].}
\renewcommand{\arraystretch}{0.5}
\label{tab:var_3}
\setlength{\tabcolsep}{3pt}
\begin{tabular}{lccc}
\toprule
Sample size $n$ & BE-(Plug-in) & BT-(Plug-in) & MB-(Plug-in) \\
\midrule
\multicolumn{4}{c}{Panel A: Time-varying MDP} \\
\midrule
50   & 0.148 [0.062, 0.341] & 0.308 [0.135, 0.527] & 0.064 [0.029, 0.098] \\
100  & 0.033 [0.015, 0.057] & 0.057 [0.035, 0.085] & 0.016 [0.006, 0.030] \\
200  & 0.009 [0.005, 0.015] & 0.012 [0.006, 0.023] & 0.004 [0.003, 0.010] \\
500  & 0.002 [0.001, 0.003] & 0.002 [0.001, 0.003] & 0.002 [0.001, 0.003] \\
1000 & 0.001 [0.000, 0.001] & 0.001 [0.000, 0.001] & 0.001 [0.000, 0.001] \\
\midrule
\multicolumn{4}{c}{Panel B: Cliff-walking environment} \\
\midrule
50   & 552.9 [551.8, 554.1] & 130294.6 [57771.7, 283078.6] & 554.7 [445.6, 555.6] \\
100  & 39.72 [39.30, 40.14]    & 15409.64 [4213.79, 33701.15]   & 40.42 [39.89, 40.72] \\
200  & 4.211 [4.027, 4.364]       & 324.875 [4.166, 2510.311]         & 4.362 [4.238, 4.483] \\
500  & 0.051 [0.032, 0.088]       & 0.052 [0.023, 0.083]              & 0.053 [0.027, 0.084] \\
1000 & 0.021 [0.008, 0.035]       & 0.024 [0.011, 0.039]              & 0.028 [0.016, 0.042] \\
\bottomrule
\end{tabular}
\end{table}

\section{Conclusions}\label{Sec:5}

This paper studies uncertainty quantification for offline policy evaluation using a model-based bootstrap (MB) approach, with guarantees of asymptotic distributional consistency. Extensive simulations demonstrate that the proposed MB procedure accurately captures the sampling distribution of the policy value estimator, yields confidence intervals with reliable coverage and competitive tightness, and provides accurate variance estimation.

The simulation studies conducted in two representative tabular RL environments lead to the following conclusions: (1) both the MC and Plug-in estimate error distributions obtained by MB can correctly characterize their true distributions in both on-policy and off-policy scenarios; (2) in most cases, the confidence intervals constructed by BE and MB exhibit similar behavior, with empirical coverage converging to the nominal level as the sample size increases; however, in small-sample and off-policy regimes, MB consistently outperforms BE; (3) variance estimates of MC and Plug-in estimate based on MB are generally more accurate than those based on BE and BT, particularly when the sample size is limited; (4) both confidence interval construction and variance estimation based on BT perform substantially worse than BE- and MB-based methods, with deficiencies being most pronounced in small-sample settings. 

One limitation of the current method is that it relies on the support-overlap assumption, requiring the support of the target policy to be contained in that of the behavior policy. Extending it to settings with limited support overlap is an important direction for future work. Overall, the model-based bootstrap resampling method holds promise for uncertainty quantification in offline policy evaluation scenarios. Its weaker requirements on data formats make it more applicable to real-world applications.


\bibhang=1.7pc
\bibsep=2pt
\fontsize{9}{14pt plus.8pt minus .6pt}\selectfont
\renewcommand\bibname{\large \bf References}
\expandafter\ifx\csname
natexlab\endcsname\relax\def\natexlab#1{#1}\fi
\expandafter\ifx\csname url\endcsname\relax
\def\url#1{\texttt{#1}}\fi
\expandafter\ifx\csname urlprefix\endcsname\relax\def\urlprefix{URL}\fi

\end{document}